\documentclass[10pt,twocolumn,letterpaper]{article}

\usepackage[pagenumbers]{cvpr} 

\usepackage{graphicx}
\usepackage{amsmath}
\usepackage{amssymb}
\usepackage{booktabs}

\usepackage[ruled,linesnumbered]{algorithm2e}
\usepackage{color}
\usepackage[normalem]{ulem}
\usepackage{latexsym}
\usepackage{bm}
\usepackage{comment}

\usepackage[pagebackref,breaklinks,colorlinks]{hyperref}

\newcommand\XP[1]{\textcolor{black}{#1}}

\newcommand\ZH[1]{\textcolor{black}{#1}}

\newcommand\emmm[1]{\textcolor{black}{#1}}

\newcommand\ourmodel{\textcolor{black}{Multifaceted Attention Network }}
\newcommand\ourmodelinend{\textcolor{black}{Multifaceted Attention Network}}
\usepackage[capitalize]{cleveref}
\crefname{section}{Sec.}{Secs.}
\Crefname{section}{Section}{Sections}
\Crefname{table}{Table}{Tables}
\crefname{table}{Tab.}{Tabs.}

\def\cvprPaperID{3876} 
\def\confName{CVPR}
\def\confYear{2022}

\begin{document}

\title{Boosting Crowd Counting via Multifaceted Attention\thanks{This is the pre-print version of a CVPR22 paper.}}

\author{Hui LIN$^1$,
Zhiheng MA$^2$,
Rongrong JI$^3$,
Yaowei WANG$^4$,
Xiaopeng HONG$^{5,4,1}$\thanks{Corresponding author.}\\
\textsuperscript{\rm 1} Xi'an Jiaotong University,\\
\textsuperscript{\rm 2} Shenzhen Institute of Advanced Technology, Chinese Academy of Science\\
\textsuperscript{\rm 3} Xiamen University\\
\textsuperscript{\rm 4} Peng Cheng Laboratory \\
\textsuperscript{\rm 5} Harbin Institute of Technology \\
{\tt\small linhuixjtu@gmail.com; zh.ma@siat.ac.cn; rrji@xmu.edu.cn;}\\ {\tt\small wangyw@pcl.ac.cn; hongxiaopeng@ieee.org}
}
\maketitle

\begin{abstract}
  This paper focuses on the challenging crowd counting task. As large-scale variations often exist within  crowd images, neither fixed-size convolution kernel of CNN nor fixed-size attention of recent vision transformers can well handle this kind of variations. To address this problem, we propose a \ourmodel (MAN) \XP{to improve transformer models in local spatial relation encoding. MAN}
   incorporates \emph{global attention} from vanilla transformer, \emph{learnable local attention}, and \emph{instance attention} into a counting model. Firstly, the local Learnable Region Attention (LRA) is proposed to assign attention exclusive for each feature location  dynamically. Secondly, we design the Local Attention Regularization to supervise the training of LRA by minimizing the deviation among the attention for different feature locations. Finally, we provide an Instance Attention mechanism to focus on the most important instances dynamically during training. Extensive experiments on four challenging crowd counting datasets namely ShanghaiTech, UCF-QNRF, JHU++, and NWPU have validated the proposed method. Code: https://github.com/LoraLinH/Boosting-Crowd-Counting-via-Multifaceted-Attention.

\end{abstract}

\begin{figure*}[t]
  \centering
  \includegraphics[width = .95\textwidth]{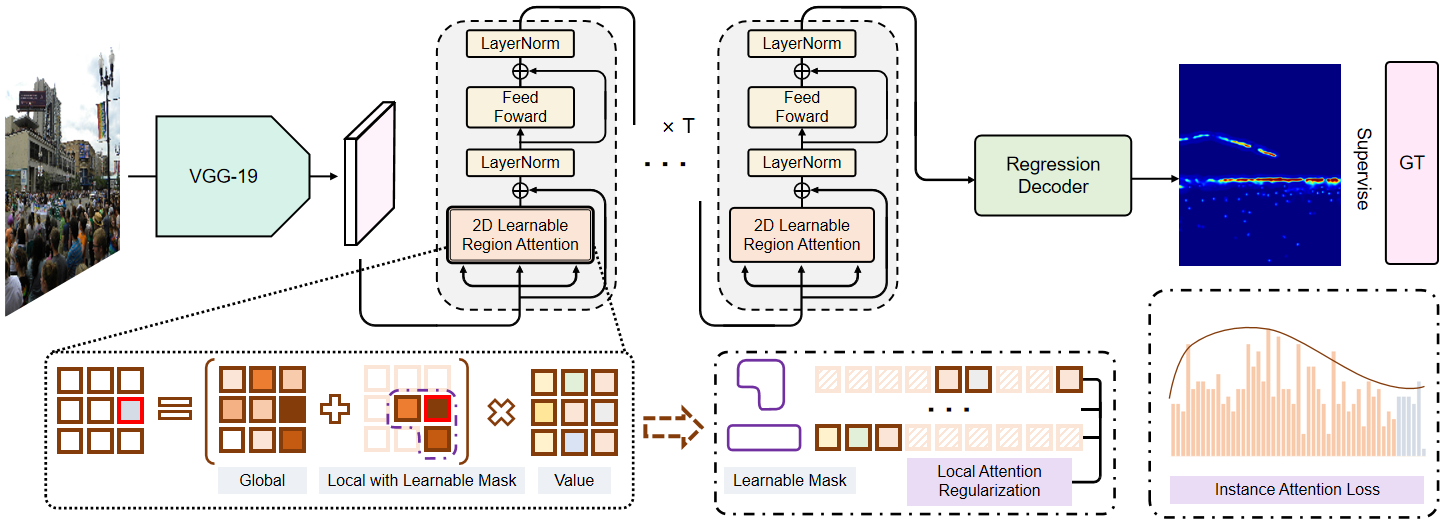}
   \caption{The framework of \ourmodelinend. \XP{A} crowd image is first fed into CNN. Then the flatten output feature map is transmitted into the transformer encoder with the \emph{Learnable Region Attention}. Finally, a regression decoder predicts the density map. \XP{\emph{Local Attention Regularization} and \emph{Instance Attention Loss} (in lilac boxes) are optimized during} the training process.}
   \label{fig:network}
\end{figure*}

\section{Introduction}
\label{sec:intro}

Crowd counting plays an essential role in congestion estimation, video surveillance, and crowd management. Especially after the outbreak of coronavirus disease (COVID-19), real-time crowd detection and counting attract more and more attention.

\XP{In recent years, typical counting methods~\cite{zhang2016single, ma2019bayesian, ma2021towards, wang2022eccnas} utilize the Convolution Neural Network (CNN) as backbone and regress density map to predict the total crowd count.}
\ZH{However, due to the wide viewing angle \XP{of cameras and the 2D perspective projection}, large-scale variations often exist in crowd images. Traditional CNNs with fixed-size convolution kernel are difficult to deal with these variations \XP{and the counting performance is severely limited. To alleviate this issue, multi-scale mechanism is designed,} such as multi-scale blobs~\cite{zeng2017multi}, pyramid networks~\cite{ma2020learning}, and multi-column networks. These methods introduce an intuitive local-structure inductive bias~\cite{xu2021vitae}, \XP{suggesting that} the respective field should be adaptive to the size of objects.}



\ZH{Lately, the blossom of Transformer models, which adopt the global self-attention mechanism, has significantly improved the performances of various natural language processing tasks. Nonetheless, \XP{it is} not until ViT\cite{dosovitskiy2020image} introduces patch-dividing as a  \XP{local-structure} inductive bias \XP{that} transformer models can compete \XP{with} and even surpass CNN models in vision tasks. The development of vision transformer \XP{suggests} that both global self-attention mechanism and local inductive bias are important for vision tasks.}

\XP{The study about transformer based crowd counting is just in its preliminary stage~\cite{liu2021swin, zhang2021multi} and  undergoes major challenges in introducing the local inductive bias to transformer models in crowded scenes. These models usually use fixed-size attention as ViT, which is limited in encoding the 2D local structure as pointed out by \cite{dosovitskiy2020image} and clearly inadequate to handle large-scale variations of crowd images. To solve this problem, in this paper, we improve both the structure and training scheme of vision transformers for crowd counting from the following three perspectives.}

\XP{Firstly, in response to such limitations in local region encoding, we propose the learnable region attention (LRA) to emphasize the local context. Different from previous vision transformers that adopt fixed patch division schemes, LRA can flexibly determine which local region it should pay attention to for each feature location. As a result, the local attention module provides an efficient way of extracting the most relevant local information  against the scale changes. Moreover, it further disengages from the dependence on the position embedding module of ViT, which has been proven inefficient in encoding local space relations~\cite{dosovitskiy2020image}.}

\XP{Secondly, we propose an efficient Local Attention Regularization (LAR) method to regularize the training of the LRA module. Inspired by the recent finding of human behaviors~\cite{collegio2019attention} that people often allocate similar attention resources to objects with similar real sizes regardless of their sizes in 2D images, we require the allocated attention \emph{w.r.t.} each feature location to be similar. Based on this understanding, we design LAR to optimize the distribution of local attention by penalizing the deviation among them. LAR enforces the span of visual attention to be small on crowd area, and vice versa, for balanced and efficient allocations of attention.}

Finally, \XP{we make an attempt to apply the attention mechanism to the instance (i.e., the point annotations) level in images and propose the \emph{Instance Attention} module. As the point annotations as provided in popular crowd benchmarks are spare and can only occupy a very small portion of the entire human heads, there are unavoidable annotation errors. To alleviate this issue, we use \emph{Instance Attention} to focus on the most important instances dynamically during training.}

In summary, we propose a counting model with multifaceted attention, termed as \ourmodel (MAN), to address large-scale variations in crowd images. The contributions are further summarized as follows:
\begin{itemize}
    \item  \XP{We propose the local Learnable Region Attention to allocate an attention region exclusive for} each feature location dynamically.
    \item \XP{We design a local region attention regularization method to supervise the training of LRA.}
    \item \XP{We introduce an effective instance attention mechanism to select the most important instances dynamically during training.}
    \item \XP{We perform extensive experiments on popular datasets including ShanghaiTech, UCF-QNRF, JHU++, and NWPU, and show that the proposed method makes a solid advance in counting performance.}
\end{itemize}

\vspace{-2mm}
\section{Related Works}

\subsection{Crowd Counting}
Existing crowd counting methods can be categorized into three types: detection, regression and point supervision. Detection based methods~\cite{liu2018decidenet, liu2019point} construct detection models to predict bounding boxes for every person in the image. The final predicted count is represented by the number of boxes. However, its performance is limited by the occlusion in congested areas and the need of additional annotations.  

Regression based methods~\cite{zhang2016single, li2018csrnet} predict count by regressing to a pseudo density map generated based on point annotations. More improvements such as multi-scale mechanisms~\cite{zeng2017multi, ma2020learning, sindagi2017generating, cao2018scale}, perspective estimation~\cite{yang2020reverse, yan2019perspective} and auxiliary task~\cite{liu2020semi, zhao2019leveraging} further promote the development of crowd counting.

Recently, many works propose to avoid the inaccurate generation of pseudo maps and directly use point supervisions. BL~\cite{ma2019bayesian} designs the loss function based on Bayesian theory, calculating the deviation of expectation for each crowd. And further works~\cite{ma2021learning, lin2021direct, wang2020distribution} focus on optimal transport and measure the divergence without depending on the assumption of Gaussian distribution.

\subsection{Transformer}
The transformer~\cite{vaswani2017attention} has rapidly been used in wide range of machine learning area. ~\cite{devlin2018bert} proposes Bidirectional Encoder Representations from Transformers (BERT) to enable deep bidirectional pre-training for language representations. ~\cite{radfordimproving} makes use of transformer to achieve strong natural language understanding through generative pre-training. ~\cite{dehghani2018universal} introduces a generalization of the Transformer model which extends theoretical capabilities.

The Vision Transformer (ViT)~\cite{dosovitskiy2020image} firstly applies a transformer architecture for vision tasks and demonstrates outstanding performances. DETR~\cite{carion2020end} further boosts the efficiency of vision transformer focusing on object detection. More recently, these advancements have boosted the effective applications of transformer in various tasks. ~\cite{zheng2021rethinking, wang2021end, strudel2021segmenter} adopt transformers on instance or semantic segmentation. ~\cite{zhu2020deformable, zheng2020end, sun2021rethinking} improve accuracy and efficiency for object detection. For tracking task, great properties of transformer are also leveraged in ~\cite{chen2021transformer, wang2021transformer, sun2020transtrack}.

\subsection{Variable Attention}
The self-attention module is a key component in many deep learning models and especially in different kinds of transformers. In order to better leverage on the ability of relative information extraction, some previous works endow the attention module with the variable property. ~\cite{yoon2018dynamic} proposes flexible self-attention module which computes attention weights over words with the dynamic weight vector. Disan~\cite{shen2018disan} introduces multi-dimensional attention and directional self-attention to perform a feature-wise selection and the context-aware representations. Longformer~\cite{beltagy2020longformer} utilizes dilated sliding window attention to combine local and global information. And ~\cite{nguyen2020differentiable} enables more focused attentions by dynamic differentiable windows.

In vision tasks, Swin Transformer~\cite{liu2021swin} designs shifted attention windows with overlap to achieve cross-window connections. \XP{The study}~\cite{vaswani2021scaling} develops blocked local attention and attention downsampling to improve speed, memory usage, and accuracy. ~\cite{yang2021focal} proposes focal self-attention to capture both local and global interactions in vision transformers. 
\XP{A 2-D version of sliding window attention as Longformer~\cite{beltagy2020longformer} is introduced to achieve a linear complexity w.r.t. the number of tokens~\cite{zhang2021multi}.}

We extend the previous variable attentions to learnable one, under the premise of large scale variations in crowd images. Our proposed 2D Learnable Region Attention (LRA) breaks the constraint of traditional fixed-size local attention windows in vision tasks and is robust to scale variations.

\section{The Proposed Method}
In this section, we will {elaborate the \ourmodelinend, which consists of three major components: the Learnable Region Attention (LRA), the {Local Attention Regularization (LAR)}, and the Instance Attention Loss.}

\subsection{Framework Overview}
Figure~\ref{fig:network} presents an overview of the framework. For each image $I$, we first use VGG-19~\cite{simonyan2014very} as our backbone to extract the features $F \in \mathbb{R}^{C \times W \times H}$, where $C$, $W$, and $H$ are the channel, width and height, respectively. Then the feature map \XP{is} flattened and transmitted into transformer encoder with the proposed LRA to learn features \XP{$F^{\prime}$} against various scales. \XP{Afterwards}, a regression decoder is utilized to predict the final density map $D \in \mathbb{R}^{W^{\prime} \times H^{\prime}}$ from  $F^{\prime}$. \XP{Finally, We use an \emmm{Local Attention Regularization} dedicated to supervise the training of the} LRA module and an \emmm{Instance Attention Loss} to \XP{constrain} the training of the total network. 

\subsection{Global Attention}
Traditional transformer network~\cite{vaswani2017attention} adopts self-attention layer in the encoder. It is able to connect all pairs of input and output positions to consider the global relations of current features. It is computed by:
\begin{equation}
    Att(Q, K, V) = \mathcal{S}(\frac{(QW^Q)(KW^K)^T}{\sqrt{d}})(VW^V)\XP{,}
\end{equation}
where $\mathcal{S}$ is the softmax function and $\frac{1}{\sqrt{d}}$ is a scaling factor based on the vector dimension $d$. $W^Q, W^K, W^V \in \mathbb{R}^{d \times d}$ are weight matrices for projections. $Q, K, V$, which are derived from source features, \XP{stand} for the query, key, and value vectors, respectively. 

However, since it regards the input as a disordered sequence and indiscriminately considers all correlations among features, the global attention model is position-agnostic~\cite{d2021convit}.
Therefore, we propose a local learnable region attention to consider spatial information and enable more focus attentions.


\subsection{Region Attention}
{As fixed-size convolution kernel and predesigned attentions~\cite{liu2021swin, zhang2021multi} are insufficient to learn cross-scale spatial information}, we \XP{seek} to design a mechanism by which each feature will be learnable to attend to a most suitable local region. In specific, as a rectangular region can be identified by two vertices, we begin with a region filter mechanism to obtain the exclusive region for each position.

We first define two filter \XP{functions} \XP{of} position $\textbf{p} = \left ( x_p, y_p \right )$ where $0 \leq x_p < W, 0 \leq y_p < H$ as:
\begin{equation}
\begin{aligned}
    &fil^{bl}(\textbf{p}\ |\ \textbf{b}) = \left\{
    \begin{aligned}
    &1, &\XP{\textup{if}} \  x_b \leq x_p < W, y_b \leq y_p < H \\
    &0, &\XP{\textup{otherwise}}
    \end{aligned}
    \right.,\\
    &fil^{ur}(\textbf{p}\ |\ \textbf{u}) = \left\{
    \begin{aligned}
    &1, &\XP{\textup{if}} \  0\leq x_p \leq x_u, 0\leq y_p \leq y_u \\
    &0, &\XP{\textup{otherwise}}
    \end{aligned}
    \right..
\end{aligned}
\end{equation}

Given two predicted vertices bottom left (bl) and upper right (ur): $\textbf{b} = (x_{b}, y_{b})$, $\textbf{u} = (x_{u}, y_{u})$ for a specific feature, the filter regions for it can be calculated by:
\begin{equation}
    R^{bl} = [fil^{bl}(\textbf{p}\ |\ \textbf{b})]_{p}^{W \times H}, \quad R^{ur} = [fil^{ur}(\textbf{p}\ |\ \textbf{u})]_{p}^{W \times H}.
\end{equation}

After that, calculated by Hadamard product between two filter regions $R = R^{bl} \circ R^{ur} $, the region map $R \in \mathbb{R}^{W \times H}$ can finally be expressed as:
\begin{equation}
\label{eq:hard_R}
    R(\textbf{p}) = \left\{
    \begin{aligned}
    &1, &\XP{\textup{if}} \  x_b \leq x_p \leq x_u, y_b \leq y_p \leq y_u \\
    &0, &\XP{\textup{otherwise}}
    \end{aligned}
    \right..
\end{equation}
Especially, when adopting global attention, $R$ can be represented as an all-ones matrix.

It is worth noticing that following the above-mentioned filter mechanism, the accuracy of each exclusive region entirely depends on only two discrete points, which is not learnable and lacks flexibility. 

Therefore, to explore more on local relations and improve the effectiveness of learning ability, we redesign the region filter mechanism based on coverage probability projections.

\subsection{Learnable Region Attention}

First, given the query vector and key vector $Q, K \in \mathbb{R}^{WH \times d}$, we replace the two binary filter regions by first {introducing} two predicted coverage probability maps as follows:
\begin{equation}
\begin{aligned}
    &C^1 = \mathcal{S}((QW_1^Q)(KW_1^K)^T), \\
    &C^2 = \mathcal{S}((QW_2^Q)(KW_2^K)^T),
\end{aligned}
\end{equation}
where $W_1^Q, W_1^K, W_2^Q, W_2^K \in \mathbb{R}^{d \times d}$ are trainable parameter matrices and $C^1, C^2 \in \mathbb{R}^{WH \times WH}$. 

To obtain a 2D learnable attention map, $C^1$ and $C^2$ are reshaped to \XP{an} order-3 tensor with a size of $\mathbb{R}^{WH \times W \times H}$. For each $i\in WH$ along the first axis of $C^1$ and $C^2$, there are two corresponding probability maps $\textbf{C}^1_i, \textbf{C}^2_i \in \mathbb{R}^{W \times H}$. That is, $\textbf{C}^1_i = C^1(i,:,:)$ and $\textbf{C}^2_i = C^2(i,:,:)$. We then redesign the filter region maps by following the Cumulative Distribution Function (CDF) \XP{\emph{w.r.t.}} two different directions, namely from bottom left (bl) to upper right (ur) and opposite. More specifically, given a 2D probability function $\textbf{C}_i$, for any position $\textbf{p} = \left ( x_p, y_p \right )$ 
where $0 \leq x_p < W, 0 \leq y_p < H$, we have


\begin{equation}
\begin{aligned}
&{F_{\textup{CDF}}^{bl}(\textbf{p}\ |\ \textbf{C}_i) = \sum_{x_j \leq x_p}\sum_{y_j \leq y_p} \ \textbf{C}_i (x_j, y_j)},\\
&{F_{\textup{CDF}}^{ur}(\textbf{p}\ |\ \textbf{C}_i) = \sum_{x_j \geq x_p}\sum_{y_j \geq y_p} \ \textbf{C}_i (x_j, y_j).}
\end{aligned}
\end{equation}
{Let $\hat{R}^{bl}_i\left (\textbf{C}_i  \right ) = \left [F_{\textup{CDF}}^{bl}(\textbf{p}\ |\ \textbf{C}_i) \right ]_{p}^{W \times H}$ and $\hat{R}^{ur}_i\left (\textbf{C}_i  \right ) = \left [F_{\textup{CDF}}^{ur}(\textbf{p}\ |\ \textbf{C}_i) \right ]_{p}^{W \times H}$,} the learnable region map $\hat{R}_i \in \mathbb{R}^{W \times H}$ is given by:
\begin{equation}
    {\hat{R}_i = \hat{R}^{bl}_i\left ( \textbf{C}_i^1 \right ) \circ \hat{R}^{ur}_i\left ( \textbf{C}_i^2 \right ),}
    \label{eq:final map}
\end{equation}
where $\circ$ is the Hadamard product.

\begin{figure}[t]
  \centering
  \includegraphics[width = 0.48\textwidth]{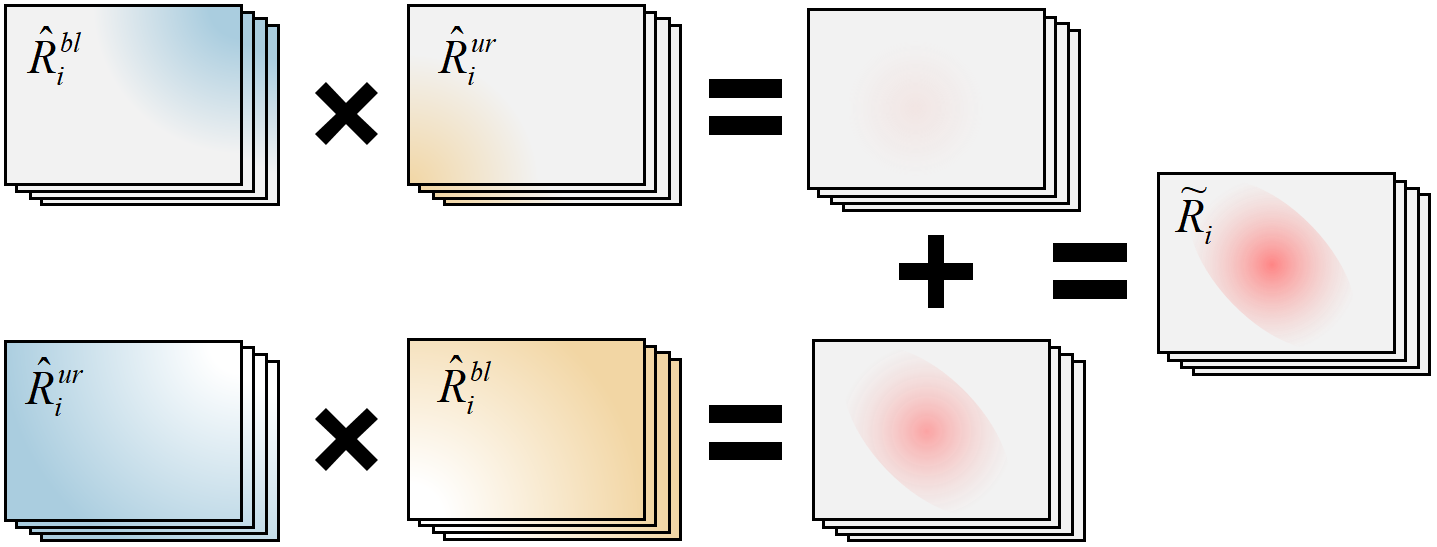}
  \caption{\XP{An example} of the learnable region filter mechanism. \XP{See texts for details.}}
    \vspace{-3mm}
   \label{fig:region2}
\end{figure}

{Nonetheless}, since we compute those two probability maps by softmax function, the two cumulative distributions $\hat{R}^{bl}_i, \hat{R}^{ur}_i$ {may} have a large number of zero values. Then the final region map, which is the product of above two regions, \XP{will} be {trivial, as illustrated by the first row of Figure~\ref{fig:region2}}. Therefore, we \XP{perform} a reverse to guarantee no matter which the cumulative direction is chosen, $\hat{R}_i\left ( \textbf{C}_i^1 \right )$ and $\hat{R}_i\left ( \textbf{C}_i^2 \right )$ will have {nontrivial} overlap, as shown by Figure~\ref{fig:region2}. The complete region map becomes:
\begin{equation}
\XP{\widetilde{R}_i = \hat{R}^{bl}_i\left ( \textbf{C}_i^1 \right ) \circ \hat{R}^{ur}_i\left ( \textbf{C}_i^2 \right ) + \hat{R}^{ur}_i\left ( \textbf{C}_i^1 \right ) \circ \hat{R}^{bl}_i\left ( \textbf{C}_i^2 \right ).}
\end{equation}

After obtaining the learnable region map $\widetilde{R}$,
we combine it into attention module for learnable local attention. With $W^Q_{loc}, W^K_{loc} \in \mathbb{R}^{d \times d}$ being specific parameter matrices of local attention, the output can be computed by:
\begin{equation}
    Att_{lra} = \mathcal{S}(\frac{(QW^Q_{loc})(KW^K_{loc})^T \circ \widetilde{R}}{\sqrt{d}})(VW^V)\XP{.}
\end{equation}
{Compared to $R$ \XP{in Eq.~\ref{eq:hard_R} which relies on} discrete vertex points, $\widetilde{R}$ is \XP{in a form of parameter arrays which are differentiable. The proposed learnable region attention mechanism is thus trainable and more flexible at determining the attentional regions.}}

Then the global attention is defined by sharing same refined value vectors with LRA:
\begin{equation}
    Att_{glb} = \mathcal{S}(\frac{(QW^Q_{glb})(KW^K_{glb})^T}{\sqrt{d}})(VW^V)\XP{.}
\end{equation}
Finally, the output of the \XP{complete attention module is a combination of} global attention and our proposed learnable region attention (LRA):
\begin{equation}
    Att = Att_{glb} + Att_{lra}\XP{.}
\end{equation}




\subsection{\XP{Local Attention Regularization}}
\label{subsec:targeting_loss}

{We take inspiration from the recent finding from the study of human behaviors that the human visual system usually pays similar attention to the  objects with similar real sizes~\cite{collegio2019attention}. To mimic such a phenomenon, we design the local region attention regularization module for supervising the training of the local learnable region attention module. The goal is to balance the distributions of local attention and penalize the deviation among the attention allocated to local regions.}



{More specifically},
given a predicted learnable region map $\widetilde{R}_i \in \mathbb{R}^{W \times H}$ and a feature map $F \in \mathbb{R}^{C \times W \times H}$, we \XP{compute} the attention-weighted features, \XP{which can be formulated as a double tensor contraction of the second and the third mode of $F$ with the first and second mode of $\widetilde{R}_i$~\cite{comon2014tensors}:}
\begin{equation}
    E_i = [F]_{\left (1,\left [ 2,3 \right ]  \right )} \XP{\star}   [\widetilde{R}_i  ]_{\left (\left [ 1,2 \right ]  \right )}.
\end{equation}
\XP{$[\cdot]_{(\cdot)}$ indicates the mode for the tensor contraction operator \XP{$\star$}~\cite{naskovska2020using} and we finally have $E_i\left ( c \right ) = \sum_{w=1}^{W} \sum_{h=1}^{H} F\left (c, w,h \right ) \cdot \widetilde{R}_i\left ( w,h \right ).$}

To keep the consistency of allocations of attention resources in each region, we regularize the local attention by minimizing the variance among weighted features:
\begin{equation}
    \mathcal{R}_{lra} = \sum_{i=0}^{WH} \  \mathcal{G}(E_i, \overline{E}),
\end{equation}
where $\overline{E}$ is the mean allocation of attention resources in all local attention regions. The \XP{deviation} penalty $\mathcal{G}$ between two weighted features is given by:
\begin{equation}
    \mathcal{G}(E_i, E_j) = 1 - \frac{E_i^{{T}} E_j}{||E_i|| \cdot ||E_j||}.
\end{equation}

\begin{figure}[t]
  \centering
  \includegraphics[width = 0.48\textwidth]{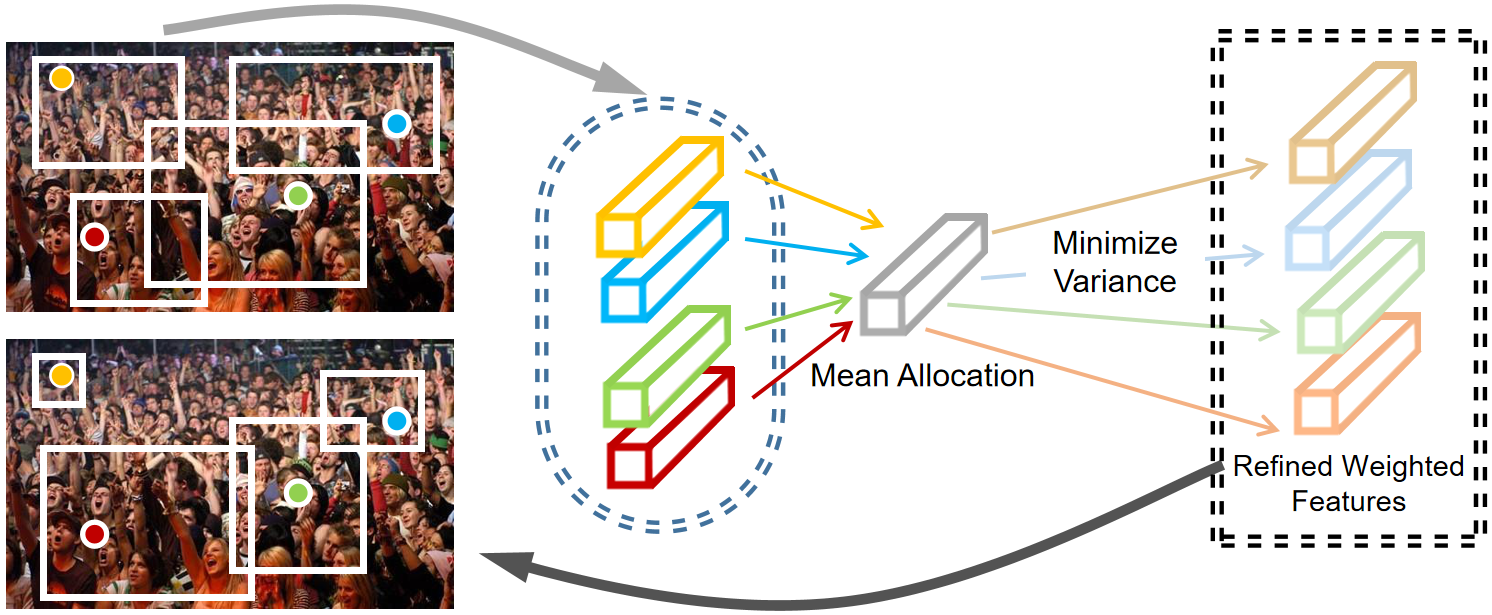}
   \caption{Overview of \XP{the} {Local Attention Regularization}. It refines the LRA by keeping the consistency of allocated attention resources in each proposed region.}
   \label{fig:feature}
\end{figure}

The scheme of Local Attention Regularization is shown in Figure~\ref{fig:feature}. 

\subsection{\XP{Instance Attention Loss}}

For \XP{optimizing} the entire network, we provide the Instance Attention Loss. As the ground truth as provided in popular crowd benchmarks is \XP{in a form of} spare point annotations and only occupies a very small portion of \XP{human heads}, this kind of human-labeled annotations inevitably exist spatial error. 

To alleviate negative influences by annotation noises, we impose a dynamic selection mechanism named Instance Attention, considering that the trained model sometimes predicts more correct signals than annotations. The mechanism is designed based on an attention mask $\mathbf{m} = [m^j]_j^N$ to select supervisions. The Instance Attention Loss is defined as:
\begin{equation}
\label{eq:ialoss}
   \mathcal{L}_{IA}=\sum_{j}^{N}\ m^j \cdot \epsilon^j,
\end{equation}
where $\mathbf{e} = [\epsilon^j]_j^N$ are deviations between predictions and labels. For example, in MSE Loss, \XP{$N$} equals to the size of density map, while in Bayesian Loss~\cite{ma2019bayesian} (BL), \XP{$N$} equals to the number of annotated points. Considering the performance and robustness, we finally choose {BL as the deviation function}:
\begin{equation}
    \XP{{\epsilon}^{j}} = |1 - \sum_{\textbf{p}} \  Prob_j(\textbf{p}) \cdot D_{\textbf{p}}|,
\end{equation}
where $j$ is $j_{th}$ annotated point. $D$ is the final predicted density map. $Prob_j(\textbf{p})$ represents for the posterior of the occurrence of the $j_{th}$ annotation given the position $\textbf{p}$.


\XP{The instance attention mask in Eq. \ref{eq:ialoss} provides a mechanism to select or weigh the instances. We regard the deviation \XP{${\epsilon}^{j}$} between predictions and labels as a kind of uncertainty of labels. If \XP{${\epsilon}^{j}$} is too large, there is a contradiction under the label of the instance. In this case, we shall reduce its importance or exclude this instance in back-propagation dynamically.}
\XP{For efficient computation, we adopt $\mathbf{m}$ as binary vectors.}
We first get the indices that sort the deviations in ascending order $\vec{\mathbf{k}} = \textup{sortID}\left ( \mathbf{e} \right )$. Then $\mathbf{m}$ is given by:

\begin{equation}
    m^j = \left\{
    \begin{aligned}
    &1, &\textup{if} \  j \in \left\{\vec{\mathbf{k}}(1),\vec{\mathbf{k}}(2),...,\vec{\mathbf{k}}([\delta N])\right\} \\
    &0, &\textup{otherwise}
    \end{aligned}
    \right.,
\end{equation}
where $\delta$ is the threshold. \XP{Clearly}, in normal cases, $\mathbf{m} = [1]^N$ and $\delta = 1.0$. In the experiments, we set $\delta = 0.9$, which means only $90\%$ annotated points with the lowest deviations from prediction will be \XP{involved in} supervision. 

Finally, \XP{the overall loss function of MAN} is defined by:
\begin{equation}
    \mathcal{L} = \mathcal{L}_{IA} + \lambda\mathcal{R}_{lra}.
\end{equation}


\def\arraystretch{1.1}
\renewcommand{\tabcolsep}{12 pt}{
\begin{table*}[t!]
\small
	\begin{center}
		\begin{tabular}{c|cc|cc|cc|cc}
			\toprule[1.5pt]
			\multicolumn{1}{c}{Dataset} & \multicolumn{2}{c}{ShanghaiTech A} &  \multicolumn{2}{c}{UCF-QNRF} &  \multicolumn{2}{c}{JHU++} & \multicolumn{2}{c}{NWPU} \\
			
			\multicolumn{1}{c}{Method} & \multicolumn{1}{c}{MAE} &
			\multicolumn{1}{c}{MSE} & \multicolumn{1}{c}{MAE} &
			\multicolumn{1}{c}{MSE} &  \multicolumn{1}{c}{MAE} &
			\multicolumn{1}{c}{MSE} &  \multicolumn{1}{c}{MAE} &
			\multicolumn{1}{c}{MSE}\\
			\hline
			\hline
			MCNN~\cite{zhang2016single}\ (CVPR 16)  & 110.2 & 173.2 & 277 & 426 & 188.9 & 483.4 & 232.5 & 714.6 \\
			CP-CNN~\cite{sindagi2017generating}\ (ICCV 17)  & 73.6 & 106.4 & - & - & - & - & - & - \\
			CSRNet~\cite{li2018csrnet}\ (CVPR 18)  & 68.2 & 115.0  &  - & - & 85.9 & 309.2 & 121.3 & 387.8 \\
			SANet~\cite{cao2018scale}\ (ECCV 18)  & 67.0  & 104.5 & - & - & 91.1 & 320.4 & 190.6 & 491.4 \\
			CA-Net~\cite{liu2019context}\ (CVPR 19)  & 61.3 & 100.0 & 107.0  & 183.0 & 100.1 & 314.0 & - & - \\
		    CG-DRCN-CC~\cite{sindagi2020jhu}\ (PAMI 20)  & 60.2  & 94.0 & 95.5 & 164.3 & 71.0 & 278.6 & - & - \\
		    DPN-IPSM~\cite{ma2020learning}\ (ACMMM 20)  & 58.1 & 91.7 & 84.7 & 147.2 & - & - & - & - \\
		    DM-Count~\cite{wang2020distribution}\ (NIPS 20) & 59.7 & 95.7 & 85.6 & 148.3 & - & - & 88.4 & 388.6 \\
		    UOT~\cite{ma2021learning}\ (AAAI 21)  & 58.1 & 95.9 & 83.3 & 142.3 & 60.5 & 252.7 & 87.8 & 387.5\\
			S3~\cite{lin2021direct}\ (IJCAI 21) & 57.0 & 96.0 & \uline{80.6} & \uline{139.8} & \uline{59.4} & \uline{244.0} & 83.5 & 346.9 \\
			P2PNet~\cite{song2021rethinking}\ (ICCV 21) & \textbf{52.7} & \textbf{85.1} & 85.3 & 154.5 & - & - & \uline{77.4} & 362.0 \\
			GL~\cite{wan2021generalized}\ (CVPR 21) & 61.3 & 95.4 & 84.3 & 147.5 & 59.9 & 259.5 & 79.3 & \uline{346.1} \\
			\hline
			BL~\cite{ma2019bayesian}\ (ICCV 19)  & 62.8 & 101.8 & 88.7 & 154.8 & 75.0 & 299.9 & 105.4 & 454.2 \\
			\textbf{MAN (Ours)} & \uline{56.8} & \uline{90.3} & \textbf{77.3} & \textbf{131.5} & \textbf{53.4} & \textbf{209.9} & \textbf{76.5} & \textbf{323.0} \\
			\bottomrule[1.5pt]
		\end{tabular}
	\end{center}
\caption{Comparisons with the state of the arts on ShanghaiTech A, UCF-QNRF, JHU-Crowd++, and NWPU. \XP{BL~\cite{ma2019bayesian} serves}  as our baseline. The best performance is shown in \textbf{bold} and the second best is shown in \uline{underlined}.}
\label{tab:performance}
\end{table*}}
\section{Experiments}

\subsection{Implement Details}

{\flushleft \textbf{Network Structure:}}
We adopt VGG-19 as our CNN backbone network which is pre-trained on ImageNet.
We refer to ~\cite{vaswani2017attention} for the structure of transformer encoder and replace the attention module by our proposed LRA. Specifically, as LRA is spatial-aware, the feature map is directly fed into the encoder without position encoding. Our regression decoder consists of an upsampling layer and three convolution layers with activation ReLU function. The kernel sizes of first two layers are $3 \times 3$ and that of last is $1 \times 1$.

{\flushleft \textbf{Training Details:}} 
We first adopt random scaling and horizontal flipping for each training image. Then we randomly crop image patches with a size of $512 \times 512$. As some images in ShanghaiTech A contain smaller resolution, the crop size for this dataset changes to $256 \times 256$. We also limit the shorter side of each image within 2048 pixels in all datasets. We use Adam algorithm~\cite{kingmaadam} with a learning rate $10^{-5}$ to optimize the parameters. We set the number of encoder layers \XP{$T$} as 4 and the loss balanced parameter $\lambda$ as 100.

\subsection{Datasets and Evaluation Metrics}
Experiments for evaluation are conducted on four largest crowd counting datasets: ShanghaiTech~\cite{zhang2016single}, UCF-QNRF~\cite{idrees2018composition}, JHU-Crowd++~\cite{sindagi2020jhu} and NWPU-CROWD~\cite{wang2020nwpu}. These datasets are described as follows:

{\flushleft \textbf{ShanghaiTech A}~\cite{zhang2016single}} contains 482 images with 244,167 annotated points. 300 images are divided for training and the remaining 182 images are for testing. Images are randomly chosen from the Internet.

{\flushleft \textbf{UCF-QNRF}~\cite{idrees2018composition}} includes 1,535 high resolution images collected from the Web, with 1.25 million annotated points. There are 1,201 images in the training set and 334 images in the testing set. UCF-QNRF has a wide range of people count with the minimum and maximum are 49 and 12,865, respectively.

{\flushleft \textbf{JHU-Crowd++}~\cite{sindagi2020jhu}} contains 4,372 images with 1.51 million annotated points. 2,772 images are chosen for training and the rest 1,600 images are for testing. The images are collected from several sources on the Internet using different keywords and specifically chosen for adverse weather conditions.

{\flushleft \textbf{NWPU-CROWD}~\cite{wang2020nwpu}} includes 5,109 images with 2.13 million annotated points. 3,109 images are divided into training set; 500 images are in validation set; and the remaining 1,500 images are in testing set. Compared with other datasets, it has the largest density range from 0 to 20,033 and contains various illumination scenes.

{\flushleft \textbf{Evaluation Metrics:}} We evaluate counting methods by two commonly used metrics: Mean Absolute Error (MAE) and Mean Squared Error (MSE). They are defined as follows:
\begin{equation}
\begin{aligned}
MAE=\frac{1}{M}\ \sum_{i=1}^{M}\big|N_i^{gt}-N_i\big|,\\ MSE=\sqrt{\frac{1}{M}\ \sum_{i=1}^{M}(N_i^{gt}-N_i)^2)}.
\end{aligned}
\end{equation}
where M is the number of sample images. $N_i^{gt}$ and $N_i$ are ground truth and estimated count of $i_{th}$ image respectively. MAE measures the accuracy of methods more and MSE measures the robustness more. The lower of both represents the better performance~\cite{zhang2016single}.

\renewcommand{\tabcolsep}{4.5 pt}{
\begin{figure*}[t!]
	\begin{center}
		\begin{tabular}{cccccccc}
			\includegraphics[width=0.11\linewidth]{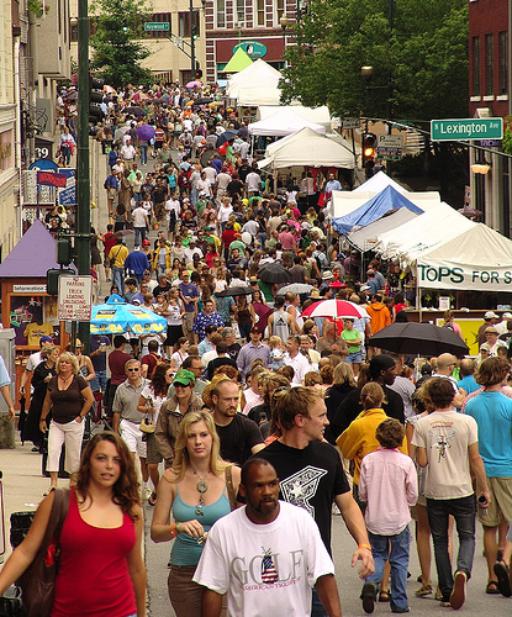}  &
			\includegraphics[width=0.11\linewidth]{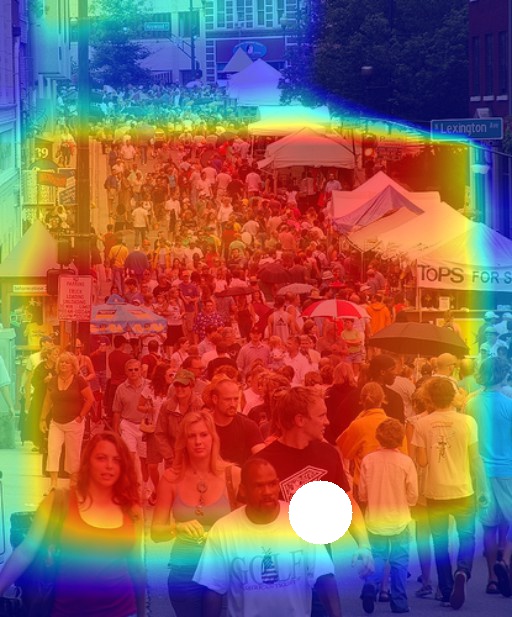}  &
			\includegraphics[width=0.11\linewidth]{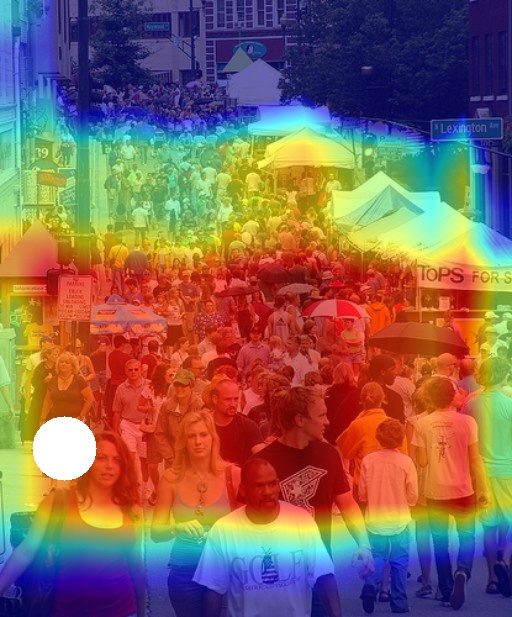}  &
			\includegraphics[width=0.11\linewidth]{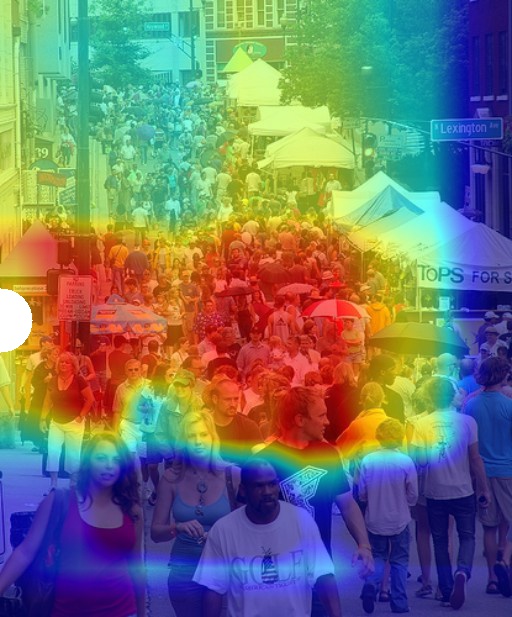}  &
			\includegraphics[width=0.11\linewidth]{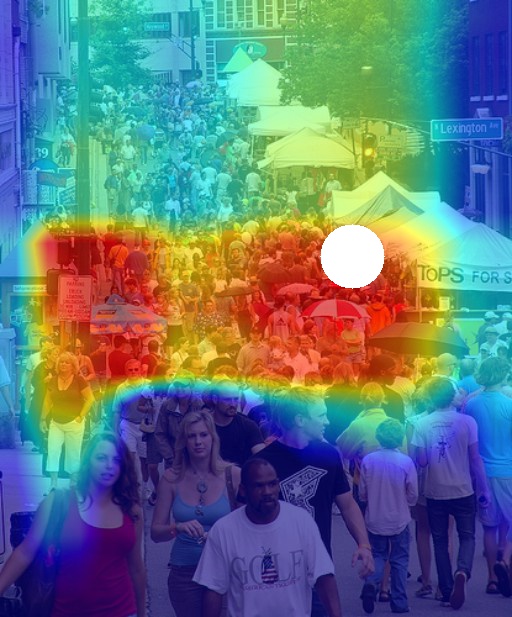}  & 
			\includegraphics[width=0.11\linewidth]{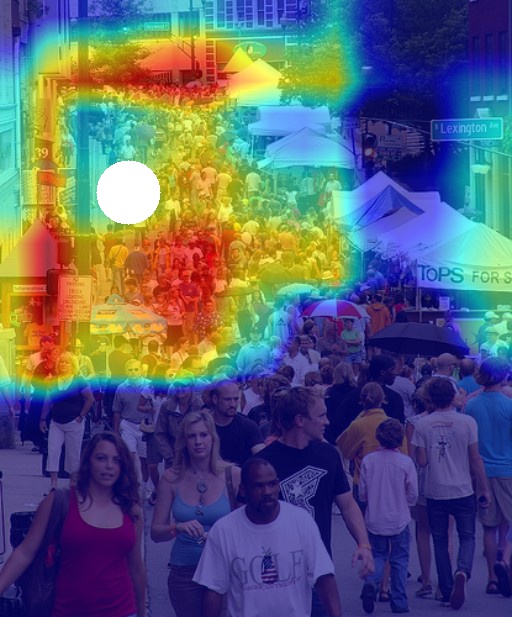}  &
			\includegraphics[width=0.11\linewidth]{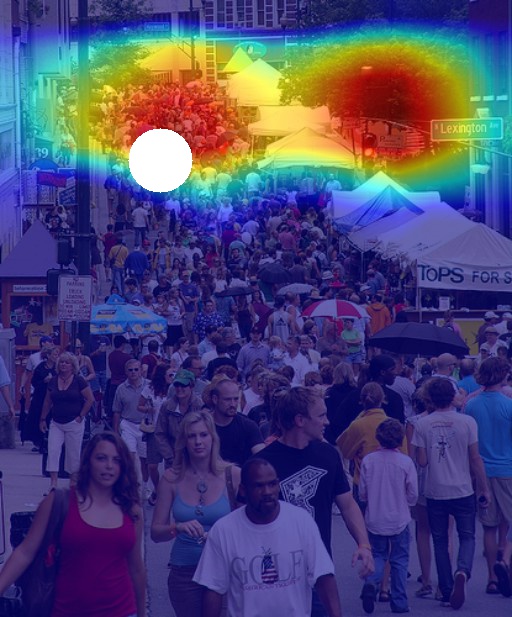} &
			\includegraphics[width=0.11\linewidth]{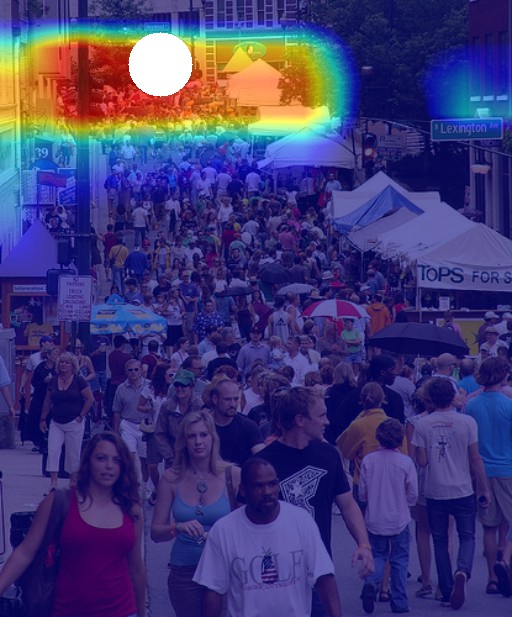}  \\
			
			\includegraphics[width=0.11\linewidth]{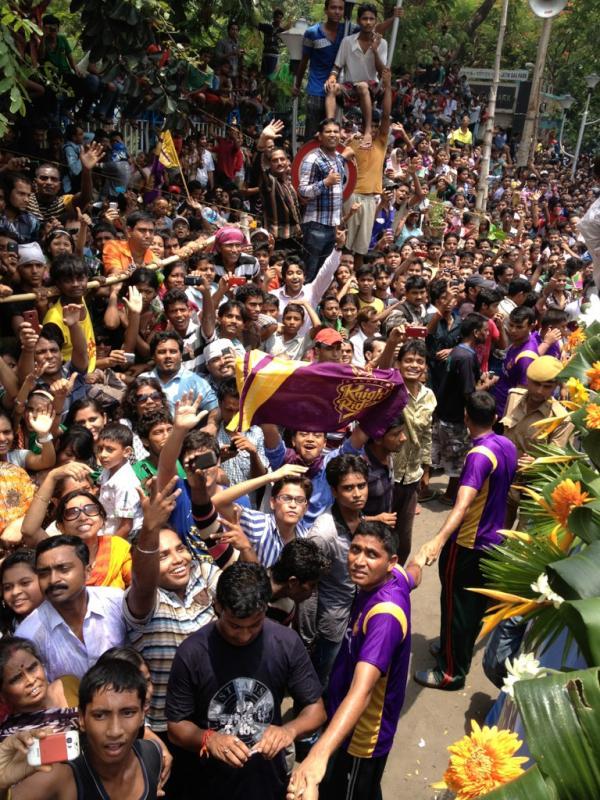}  &
			\includegraphics[width=0.11\linewidth]{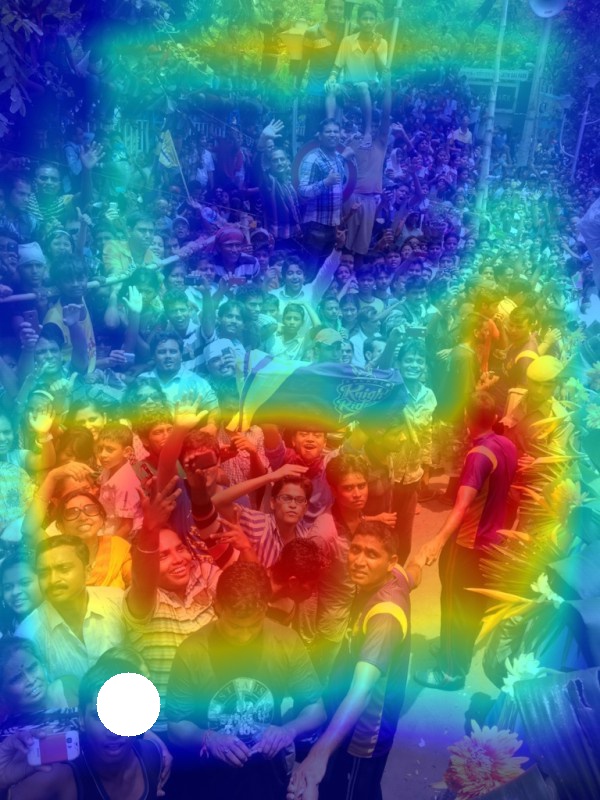}  &
			\includegraphics[width=0.11\linewidth]{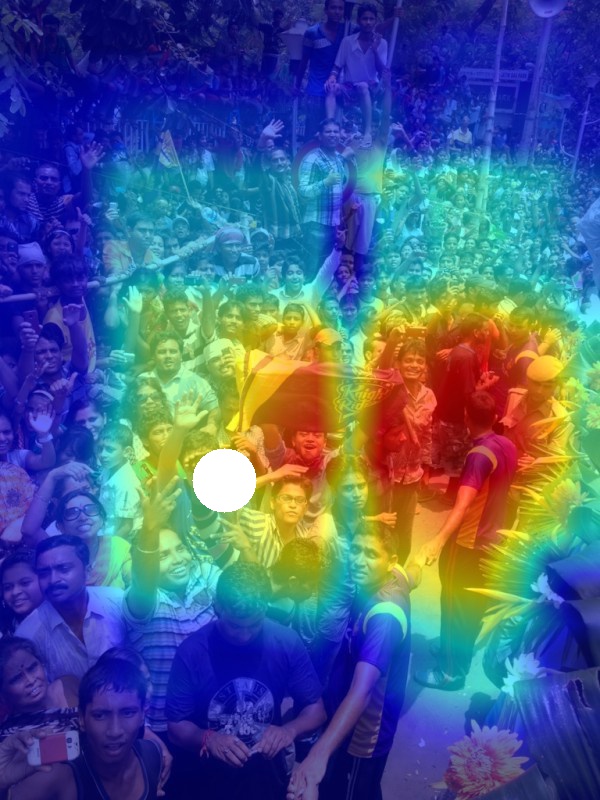}  &
			\includegraphics[width=0.11\linewidth]{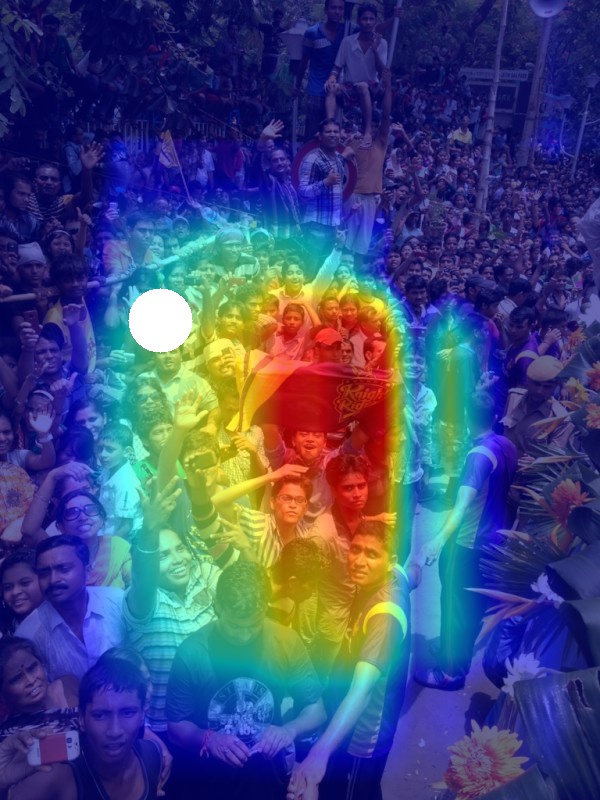}  &
			\includegraphics[width=0.11\linewidth]{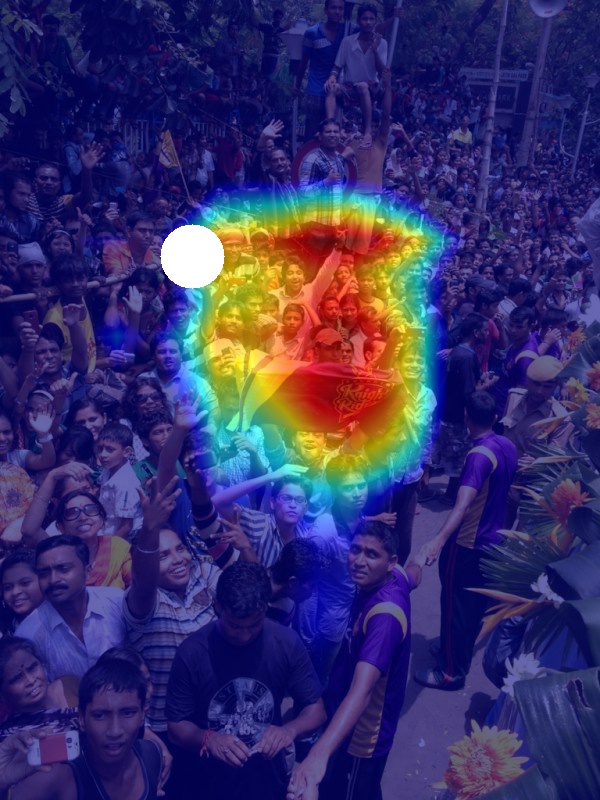}  & 
			\includegraphics[width=0.11\linewidth]{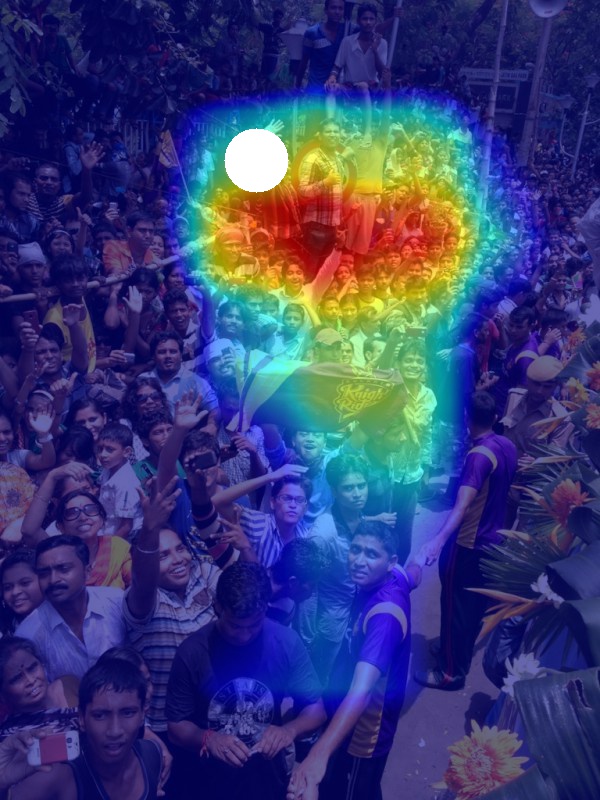}  &
			\includegraphics[width=0.11\linewidth]{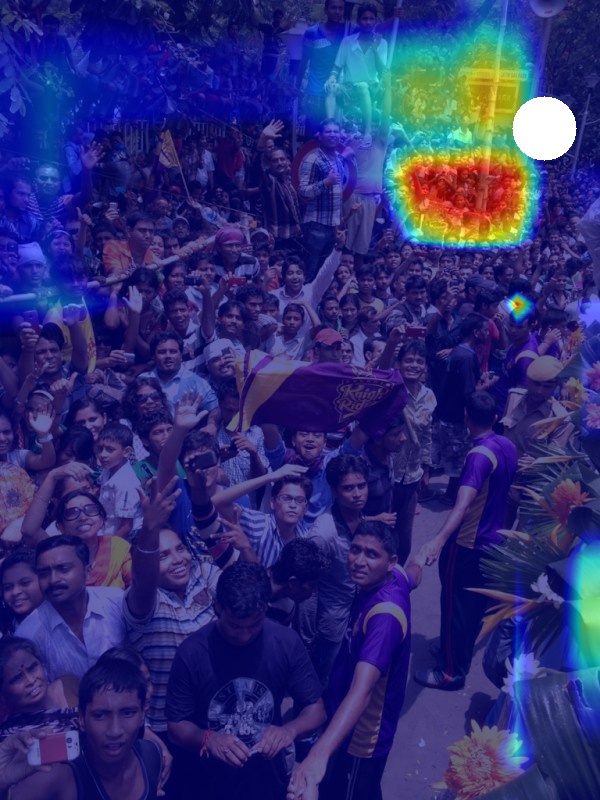} &
			\includegraphics[width=0.11\linewidth]{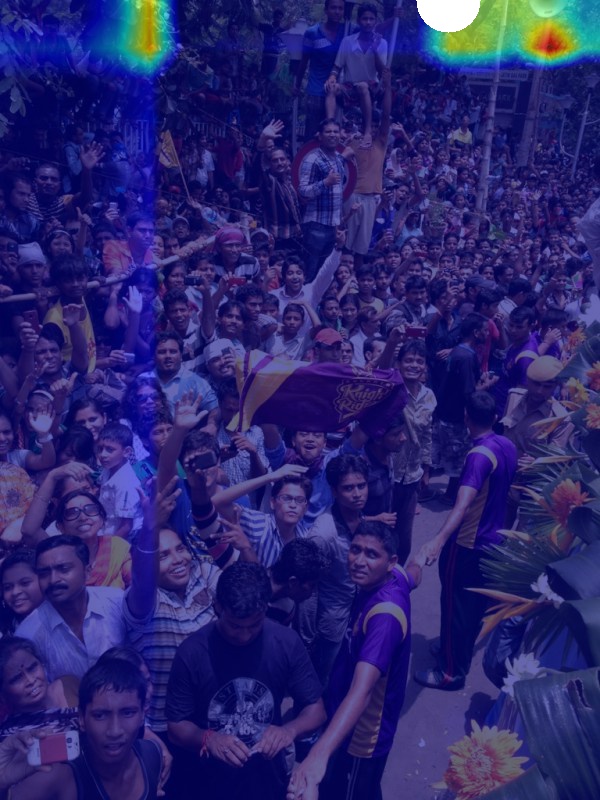}  \\

		\end{tabular}
		\caption{Visualizations of regions proposed by our Learnable Region Attention (LRA). Each white circular mark indicates the location of a feature which the mask is corresponding to. The attention region becomes gradually narrower as the scale of focus crowd is smaller.}
		\label{fig:atten}
	\end{center}
\end{figure*}}

\subsection{Comparison with state-of-the-art methods}

We evaluate our model on above four datasets and list thirteen recent state-of-the-arts methods for comparison. BL~\cite{ma2019bayesian} serves as our baseline. The quantitative results of counting accuracy are listed in Table~\ref{tab:performance}.

As the result shows, our \emmm{MAN} performs great accuracy on all the four benchmark datasets. \emmm{MAN} improves MAE and MSE values of second best method S3~\cite{lin2021direct} from 80.6 to 77.3 and from 139.8 to 131.5, respectively. On JHU++, it improves these two values from 59.4 to 53.4 and from 244.0 to 209.9, respectively.

Compared to BL, \emmm{MAN} significantly boosts its counting accuracy on all four datasets. The improvements are $9.6\%$ and $11.3\%$ for MAE and MSE on ShanghaiTech A, $12.9\%$ and $15.1\%$ on UCF-QNRF, $28.8\%$ and $30.0\%$ on JHU-Crowd++, and $27.4\%$ and $28.9\%$ on NWPU-CROWD.

Visualizations of our \emmm{MAN} are shown in Figure~\ref{fig:viz}.


\subsection{\XP{Key Issues and Discussion}}

\def\arraystretch{1.1}
\renewcommand{\tabcolsep}{7 pt}{
\begin{table}
\small
\begin{center}
\begin{tabular}{|cccc|c|c|}
\hline
Transformer & LRA & LAR & \multicolumn{1}{c}{IAL} & \multicolumn{1}{c}{MAE} & MSE\\
\hline
 & & & & 88.7 & 154.8\\
 $\checkmark$ & & & & 85.2 & 149.5\\
 & & & $\checkmark$ & 84.7 & 150.9\\
  & $\checkmark$ & & & 84.2 & 150.8\\
$\checkmark$ & & & $\checkmark$ & 83.0 & 146.2\\
$\checkmark$ & $\checkmark$ & & & 82.9 & 144.2\\
$\checkmark$ & $\checkmark$ & & $\checkmark$ & 81.5 & 137.9\\
$\checkmark$ & $\checkmark$ & $\checkmark$ & & 80.5 & 140.4\\
$\checkmark$ & $\checkmark$ & $\checkmark$ & $\checkmark$ & 77.3 & 131.5\\
\hline
\end{tabular}
\end{center}
\caption{Ablation study. Transformer is the vanilla form~\cite{vaswani2017attention}. LRA, LAR and IAL are short for the Learnable Region Attention, Local Attention Regularization and Instance Attention Loss respectively. All experiments are performed on UCF-QNRF. The full model combining all proposed modules performs best.}
\label{tab:contribution}
\end{table}}

\noindent \textbf{\XP{Ablation Studies}}
\XP{We perform the ablation study on UCF-QNRF and provide quantitative results in Table~\ref{tab:contribution}.}

We start with the baseline of BL~\cite{ma2019bayesian} \XP{and then} test the contribution of vanilla transformer encoder \cite{vaswani2017attention}. \XP{MAE and MSE are reduced} by $3.9\%$ and $3.4\%$, respectively. By adding IAL, the performance from baseline is improved by $4.5\%$ and $2.5\%$. The combination of transformer and IAL further boosts the counting accuracy.

Then, we replace the \XP{vanilla} attention module by our proposed LRA, the performance is improved by $2.7\%$ and $3.5\%$ without IAL and by $1.8\%$ and $5.7\%$ with IAL. However, it is worth noticing that when we only adopt LRA without global attention, the performance will drop, indicating both global and local information are important. Finally, when the LAR is adopted, the best performance is achieved, which boosts the counting accuracy of BL by $12.9\%$ and $15.1\%$ for MAE and MSE, respectively.

\begin{figure}[t]
  \centering
  \includegraphics[width = 0.48\textwidth]{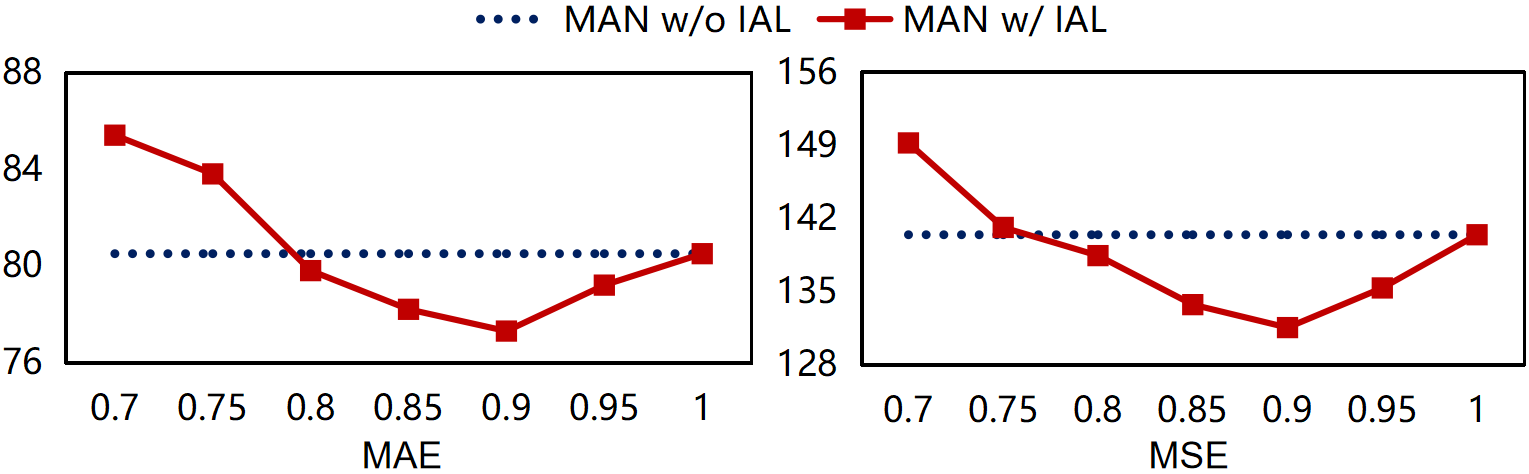}
   \caption{Effect of threshold $\delta$. The dotted line represents our network without proposed \emmm{Instance Attention Loss}. When $0.8 \leq \delta < 1$, the results are better than supervision by all annotations. }
   \label{fig:ablation1}
\end{figure}

\renewcommand{\tabcolsep}{4.5 pt}{
\begin{figure*}[t!]
	\begin{center}
		\begin{tabular}{ccccc}
			\includegraphics[height=0.13\linewidth]{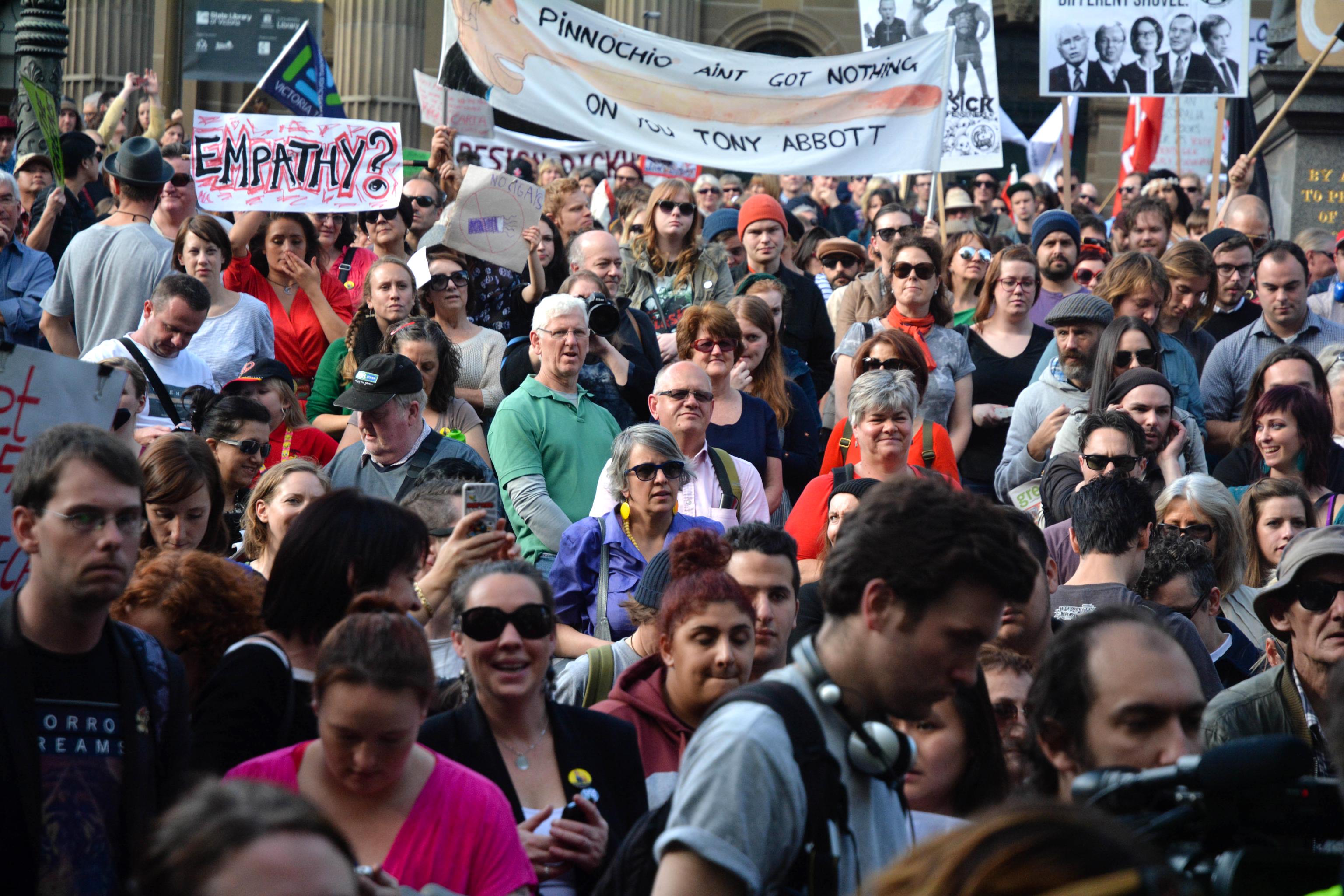}  &
			\includegraphics[height=0.13\linewidth]{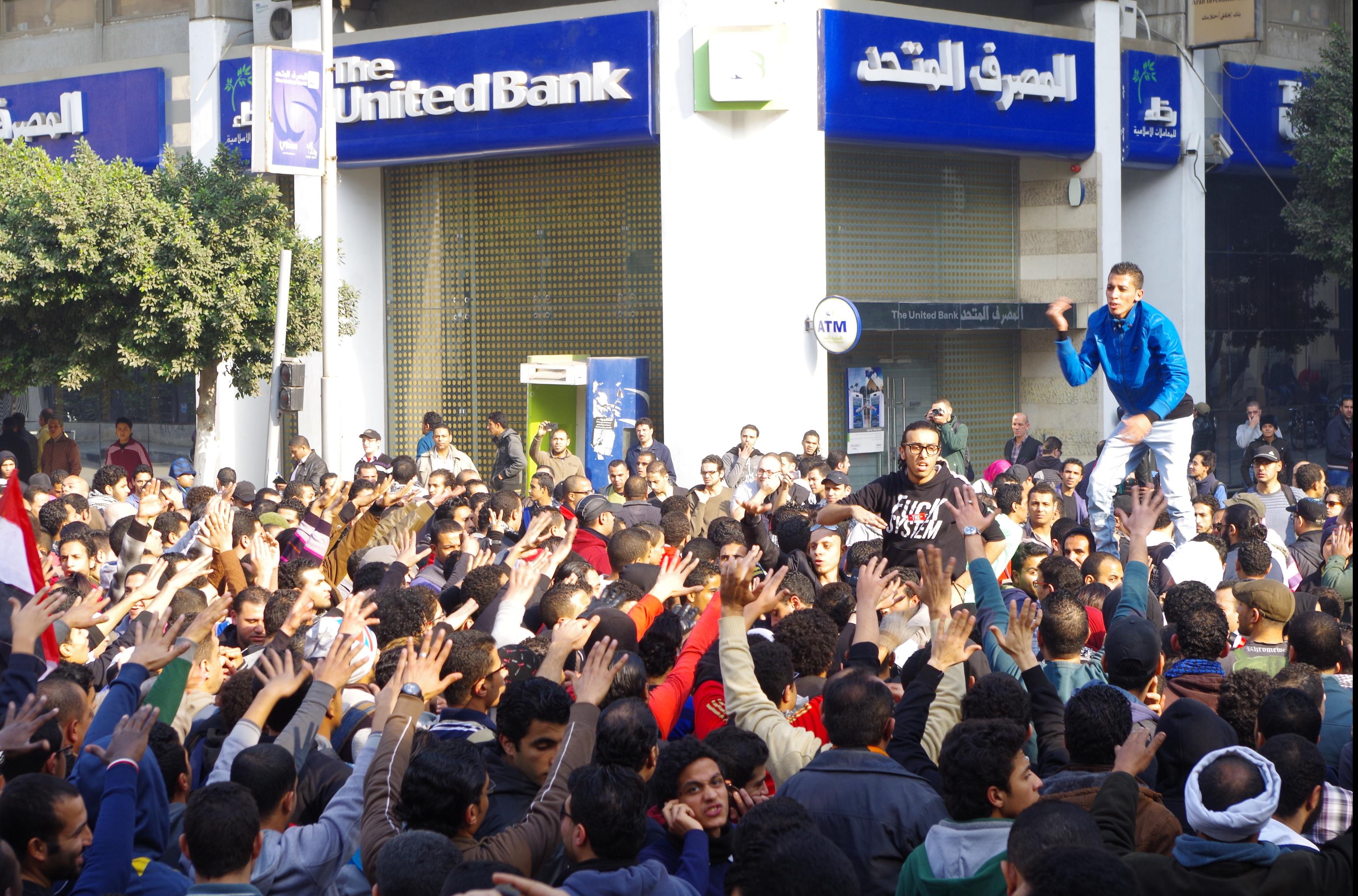}  &
			\includegraphics[height=0.13\linewidth]{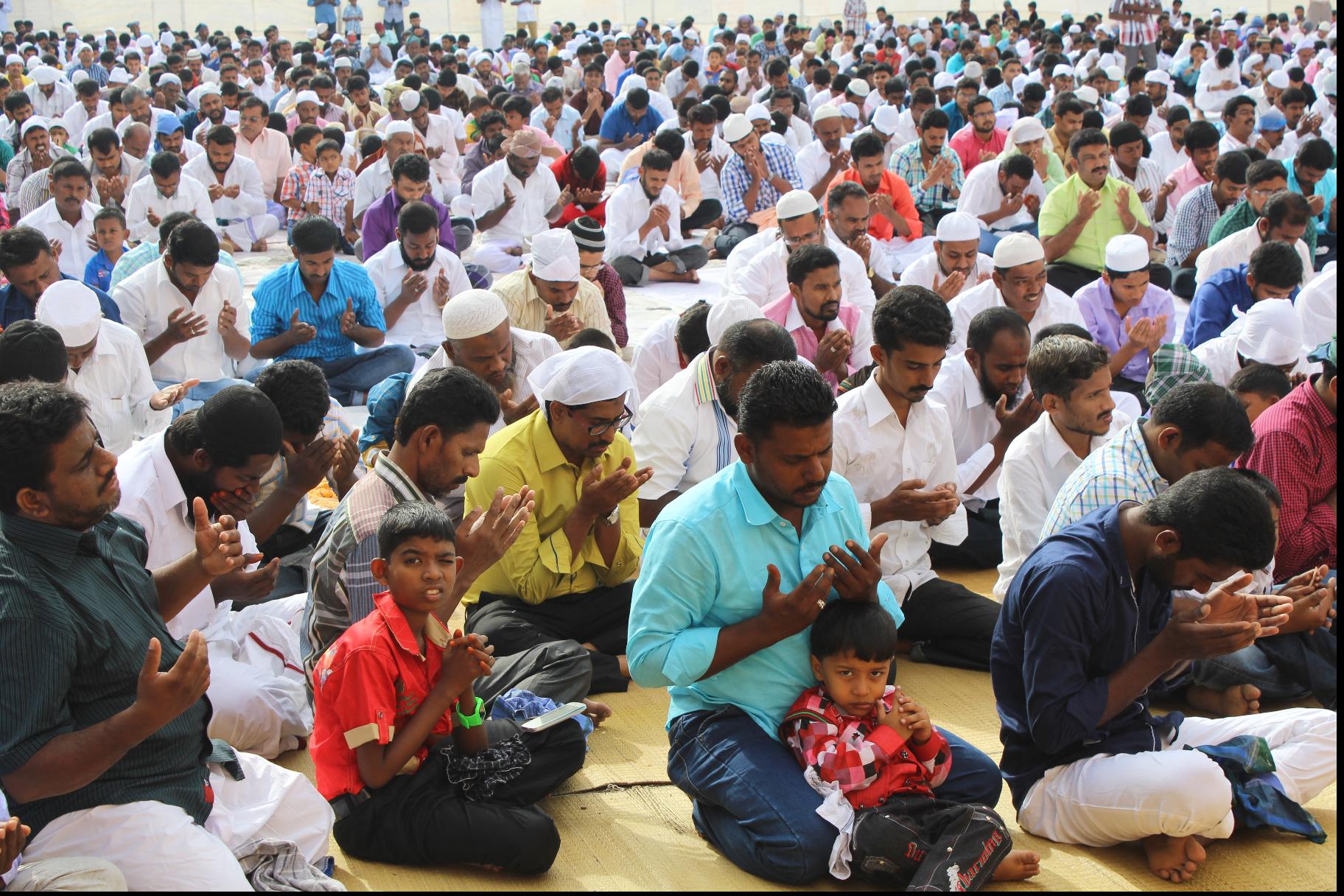}  &
			\multicolumn{2}{c}{\includegraphics[height=0.13\linewidth]{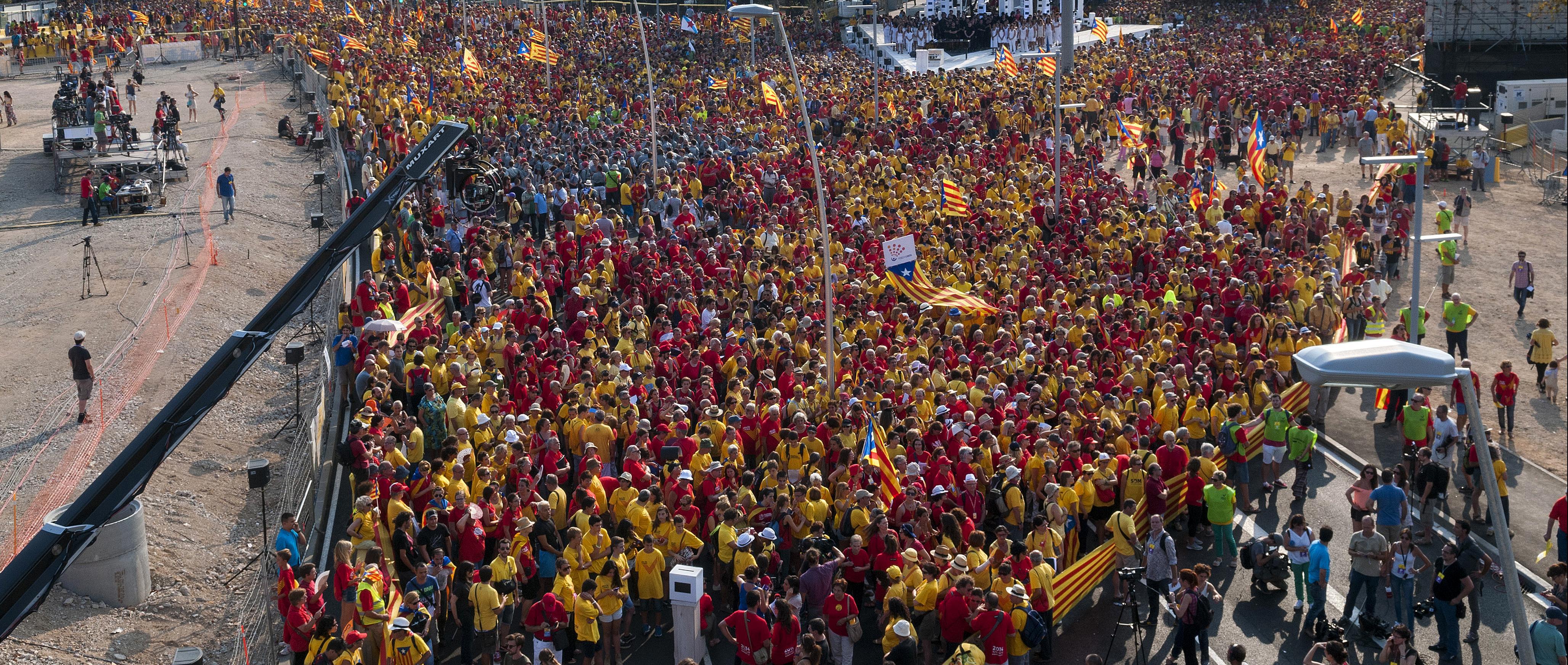}} \\
			\footnotesize{GT: 127} & \footnotesize{GT: 183} & \footnotesize{GT: 367} & \multicolumn{2}{c}{\footnotesize{GT: 4535}} \\
			
			\includegraphics[height=0.13\linewidth]{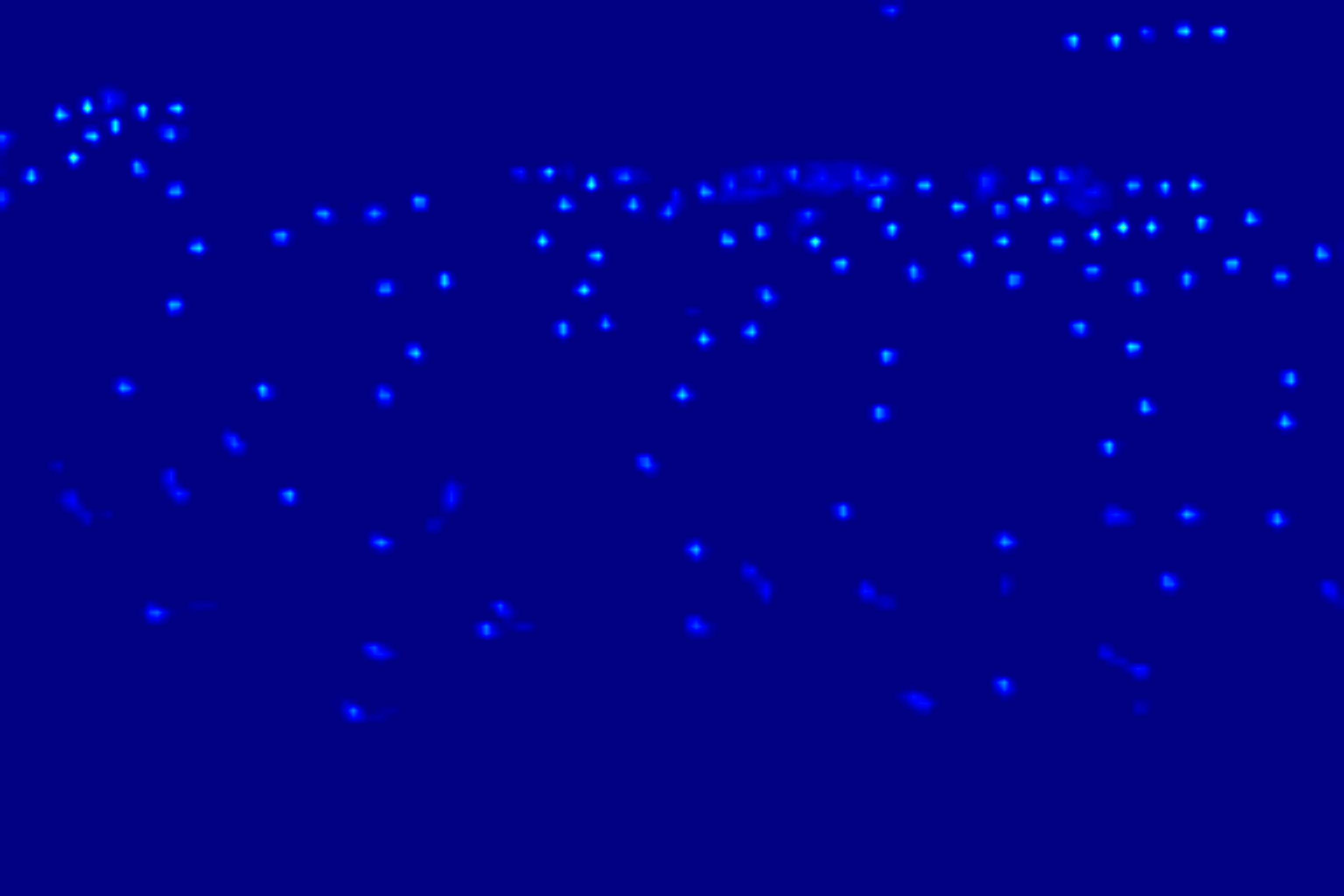}  &
			\includegraphics[height=0.13\linewidth]{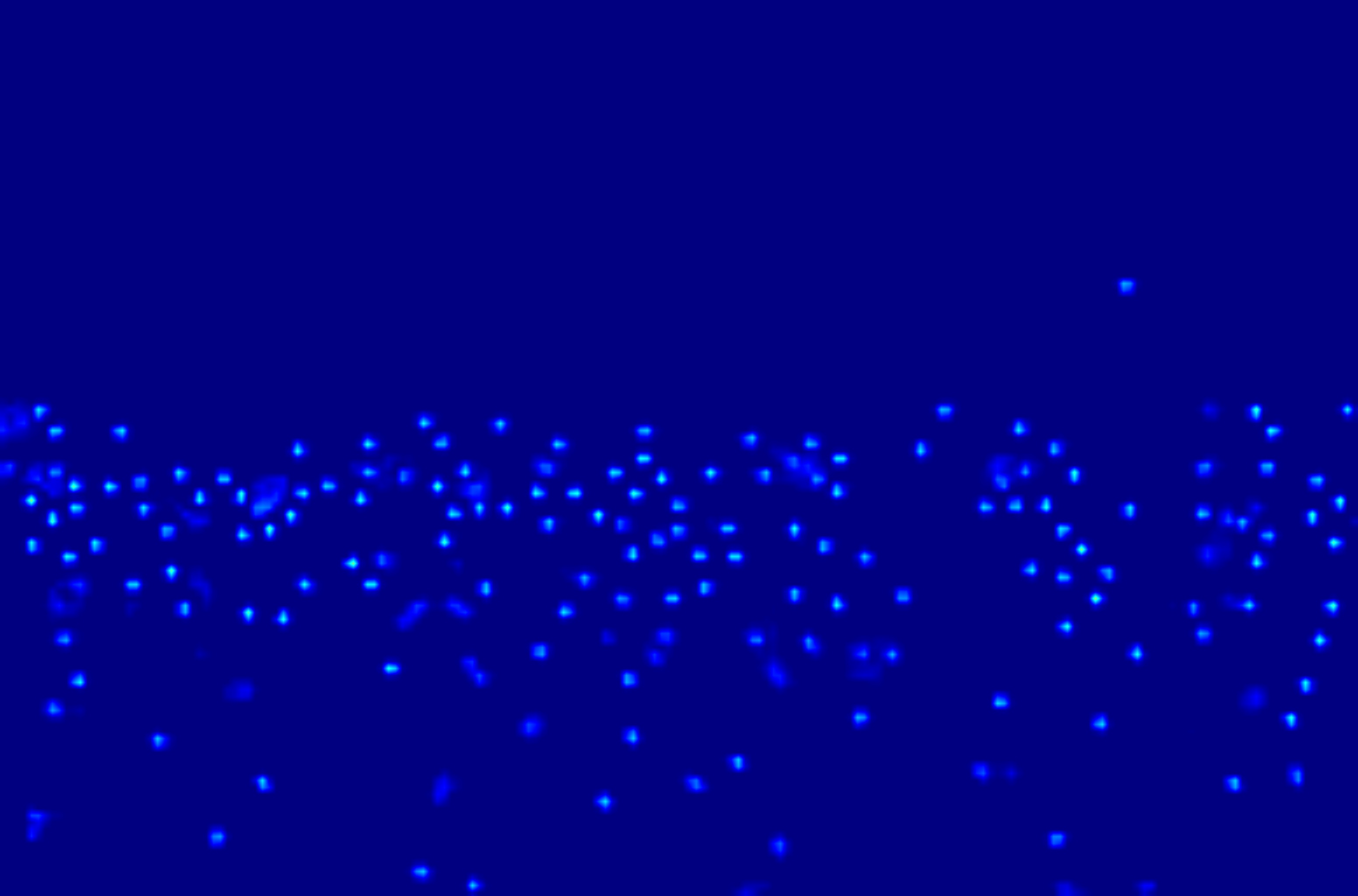}  &
			\includegraphics[height=0.13\linewidth]{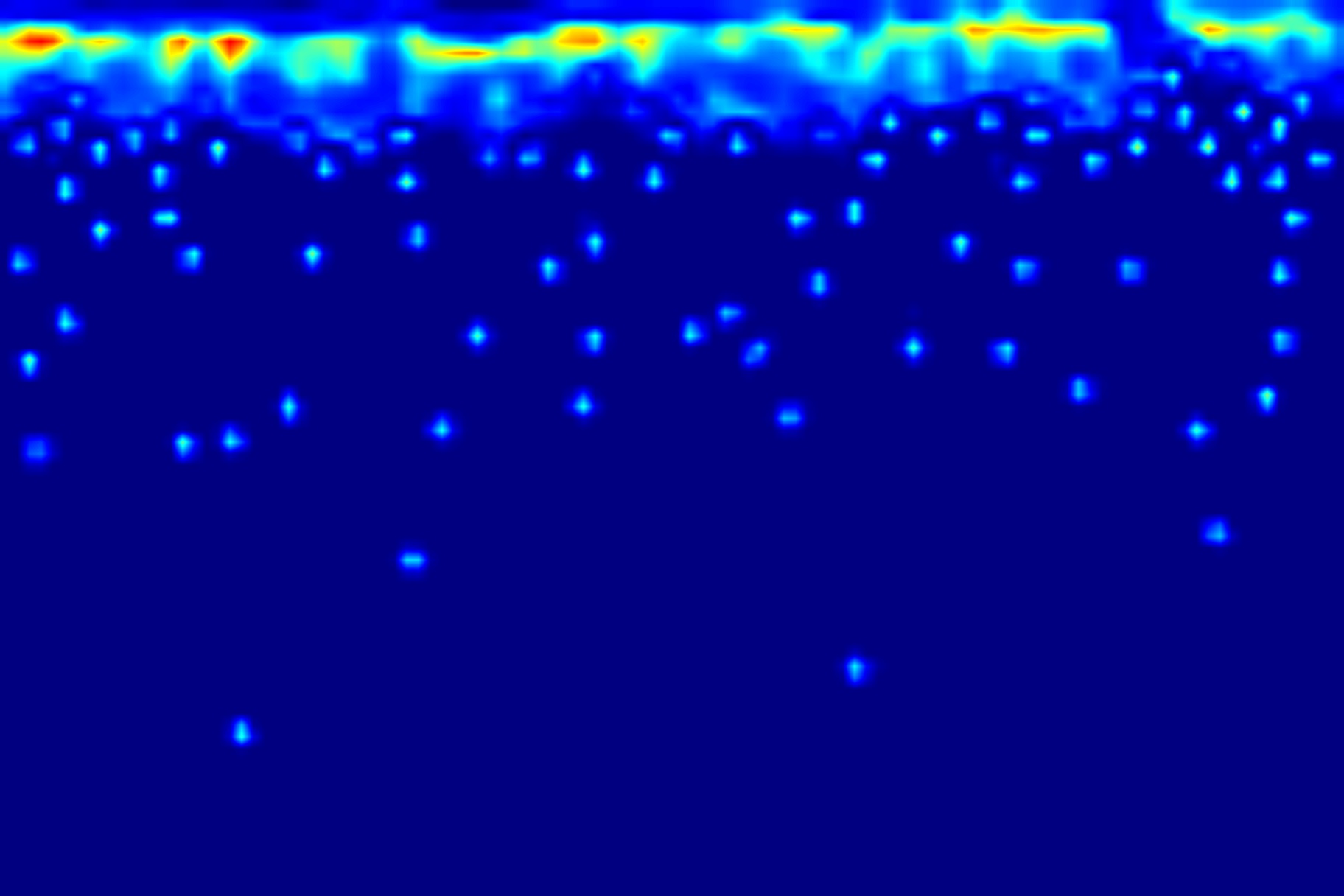}  &
			\multicolumn{2}{c}{\includegraphics[height=0.13\linewidth]{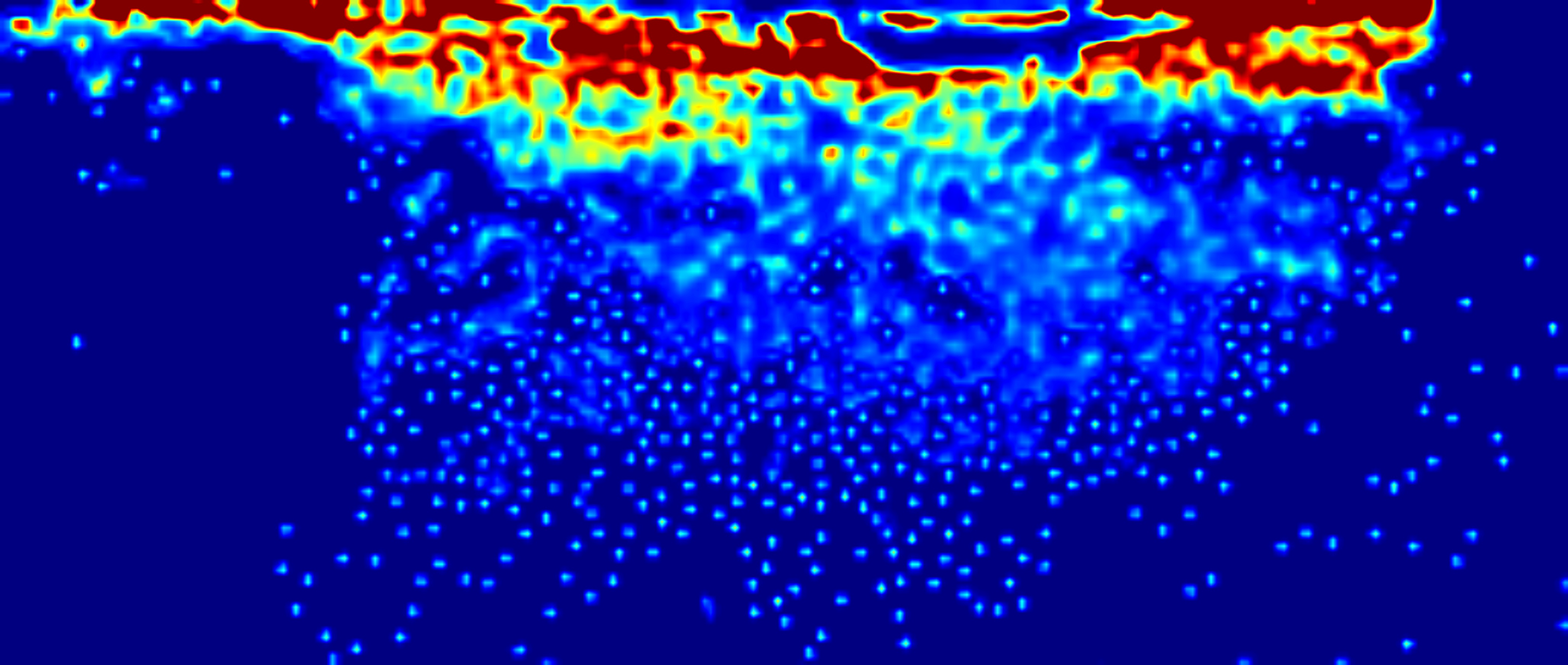}} \\

			\footnotesize{\emmm{MAN}: 125.37} & \footnotesize{\emmm{MAN}: 181.73} & \footnotesize{\emmm{MAN}: 369.01} & \multicolumn{2}{c}{\footnotesize{\emmm{MAN}: 4311.00}} \\
			
			\includegraphics[height=0.13\linewidth]{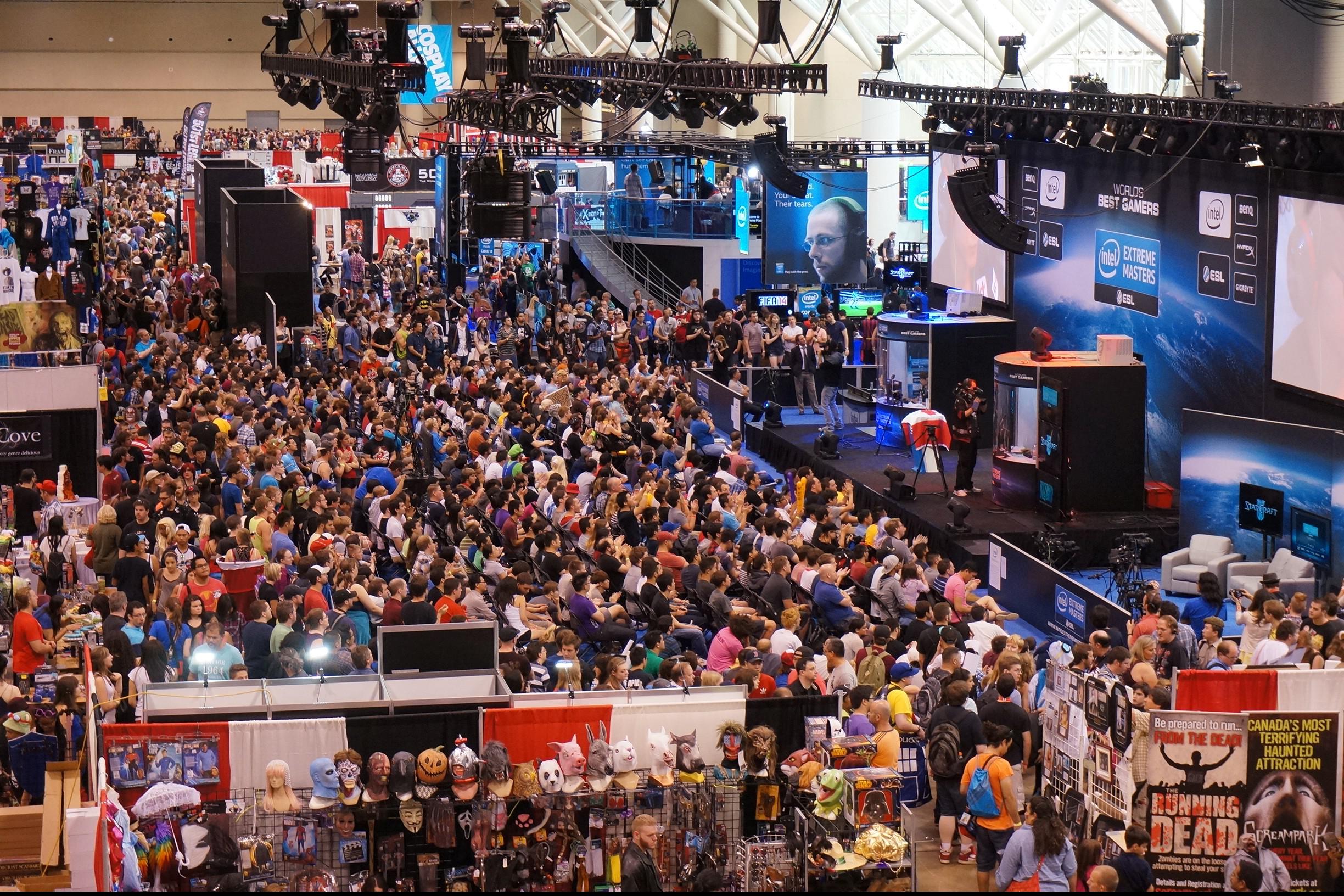}  &
			\includegraphics[height=0.13\linewidth]{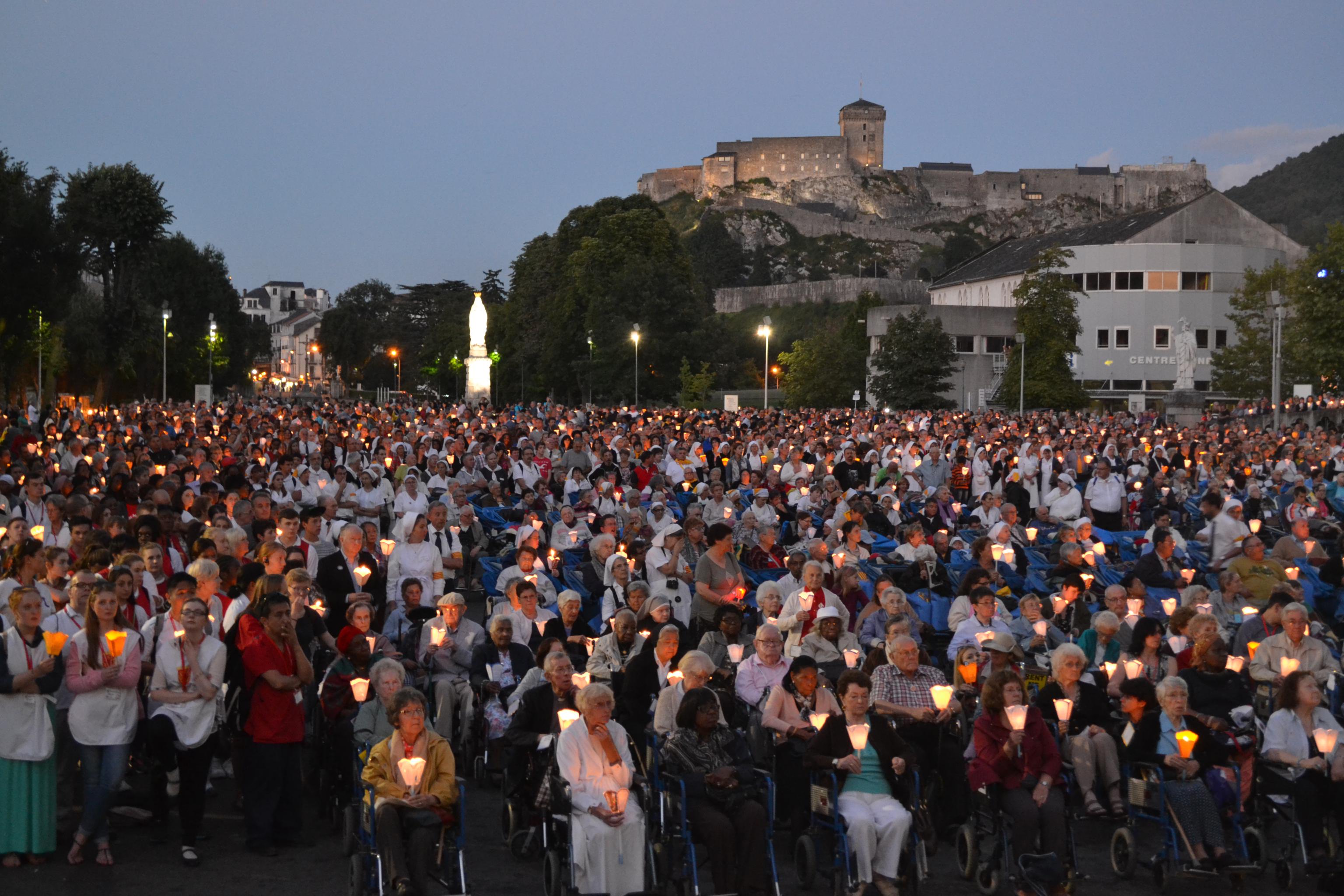}  &
			\includegraphics[height=0.13\linewidth]{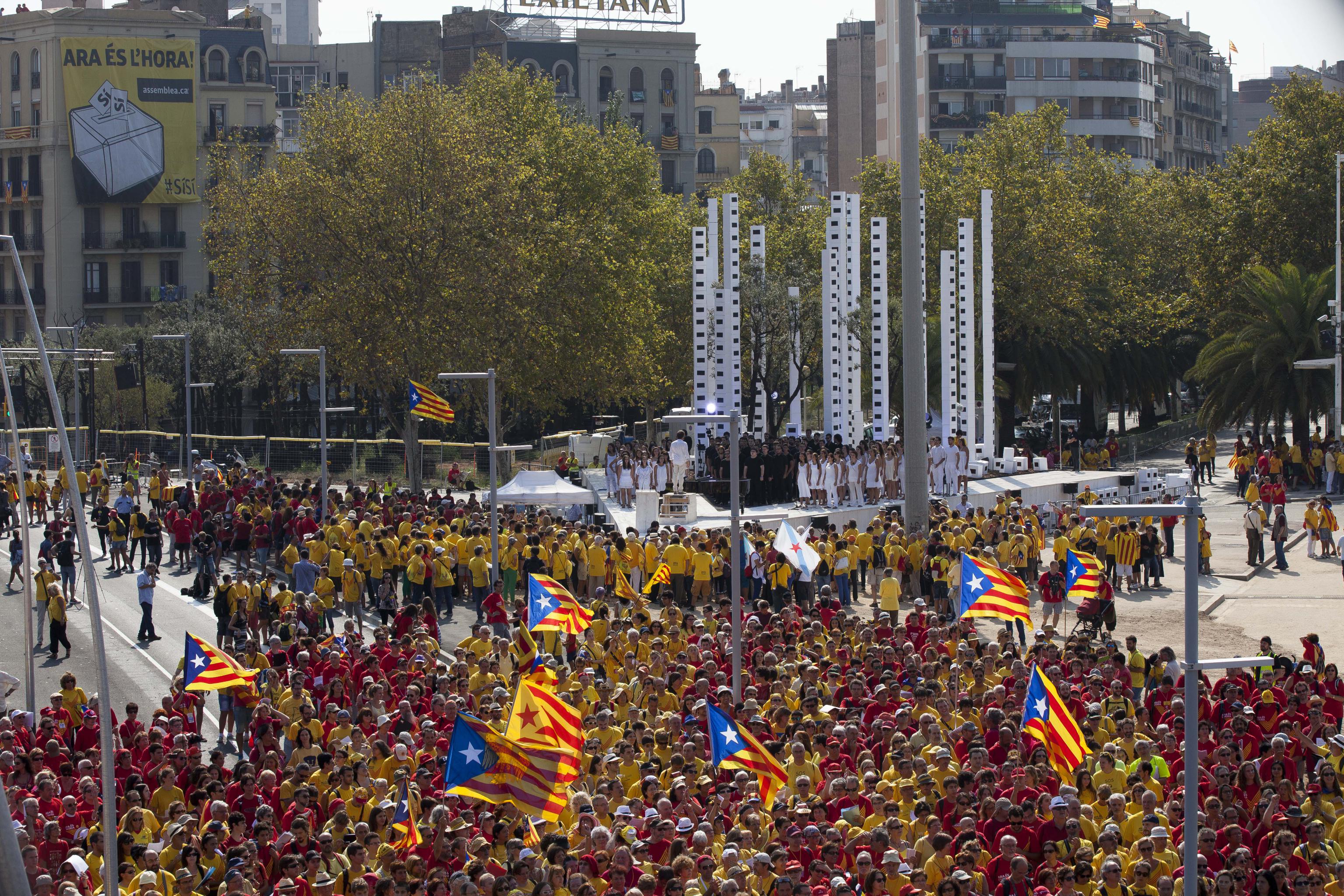}  &
			\includegraphics[height=0.13\linewidth]{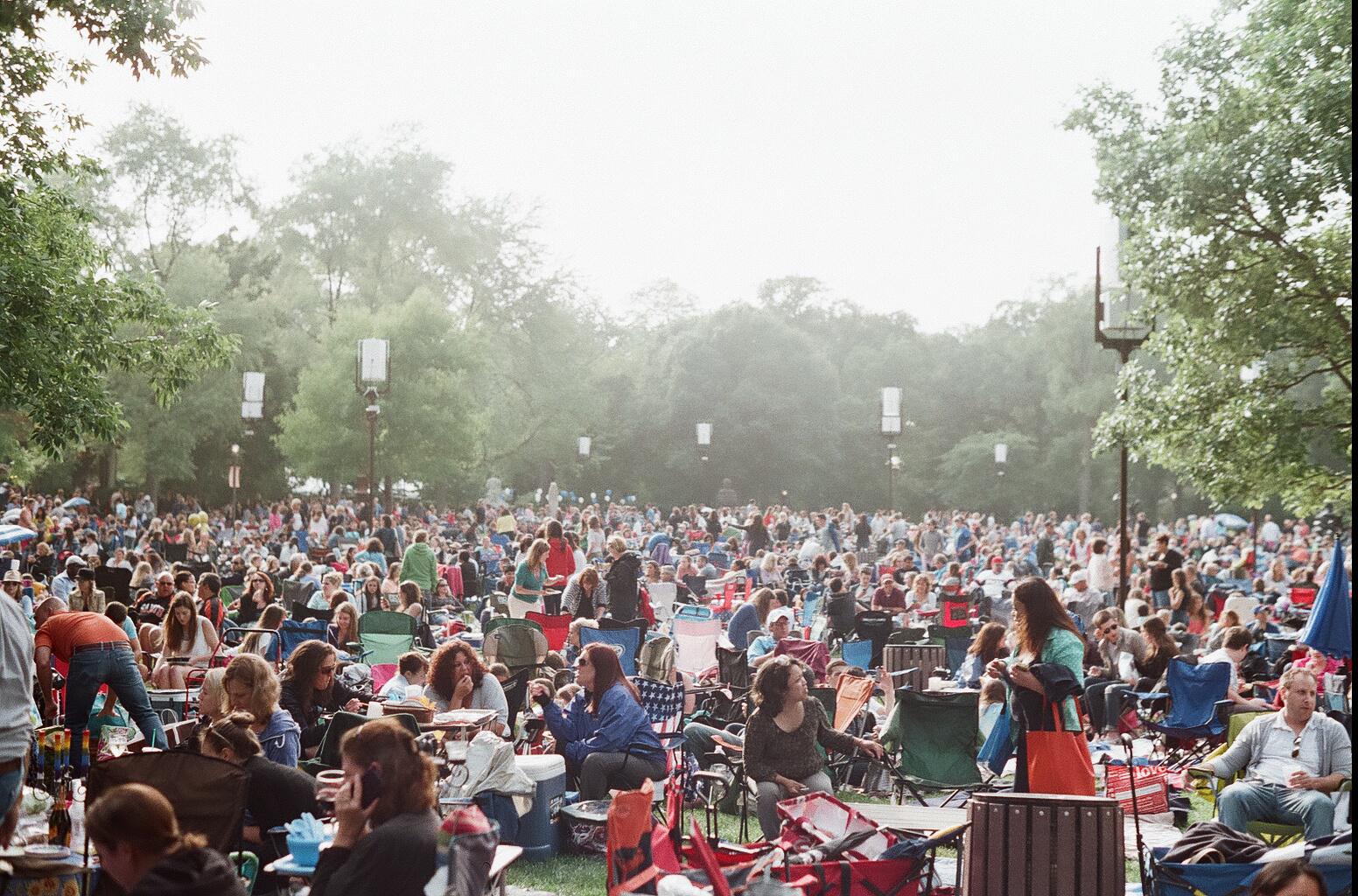} &
			\includegraphics[height=0.13\linewidth]{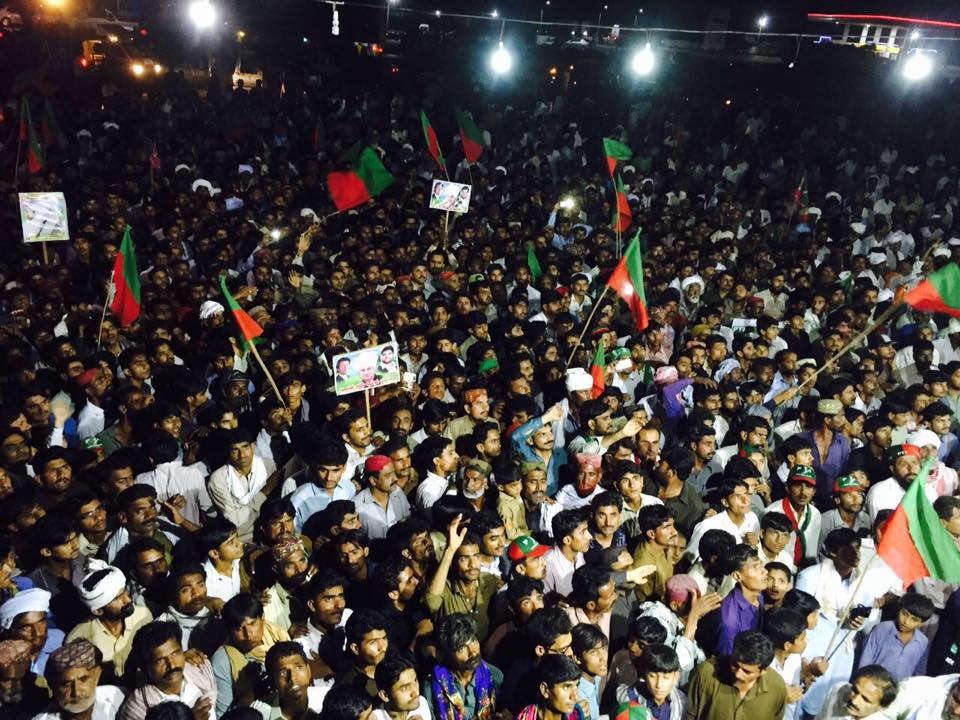} \\
			
			\footnotesize{GT: 953} & \footnotesize{GT: 1018} & \footnotesize{GT: 1211} &
			\footnotesize{GT: 405} & \footnotesize{GT: 540} \\
			
			\includegraphics[height=0.13\linewidth]{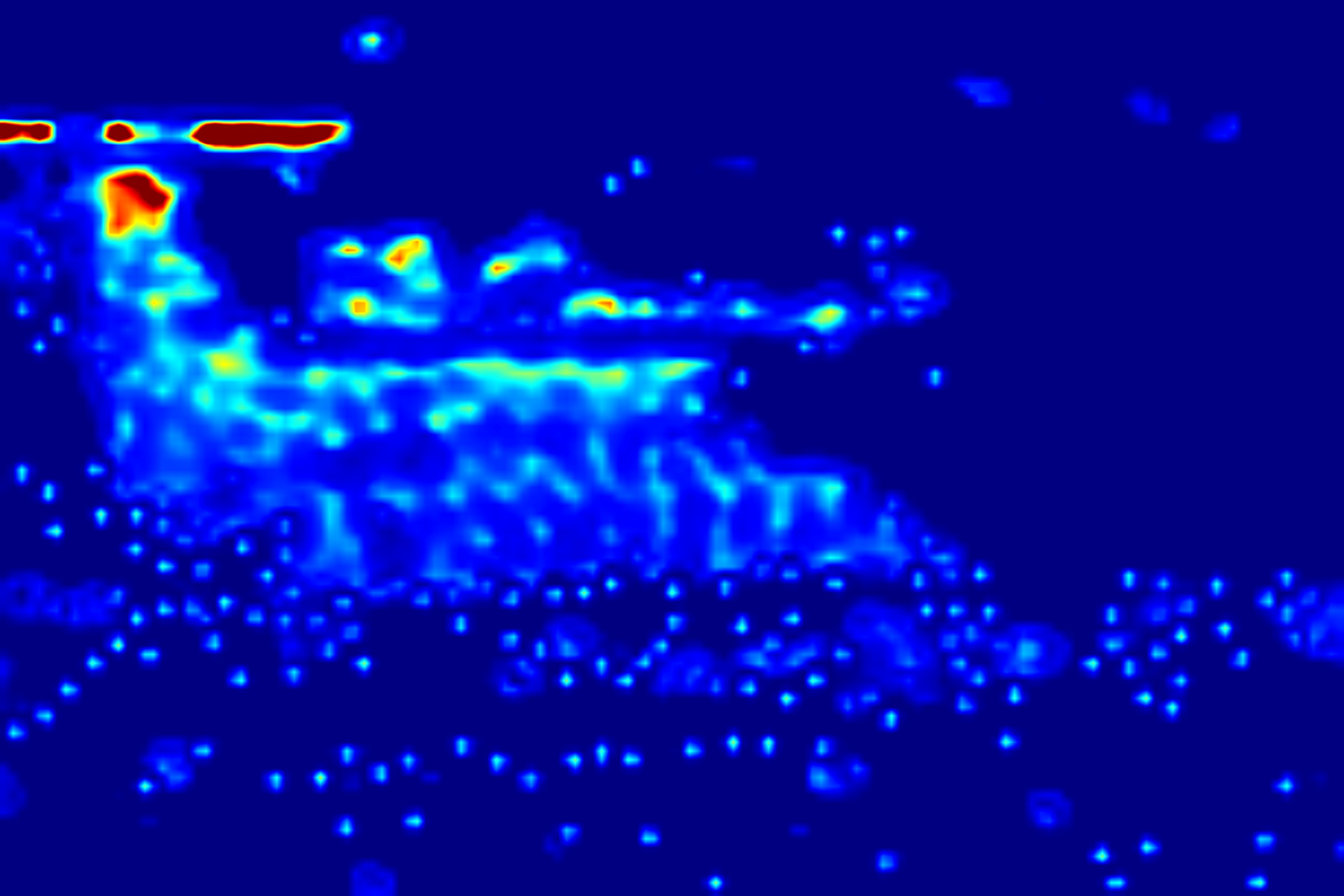}  &
			\includegraphics[height=0.13\linewidth]{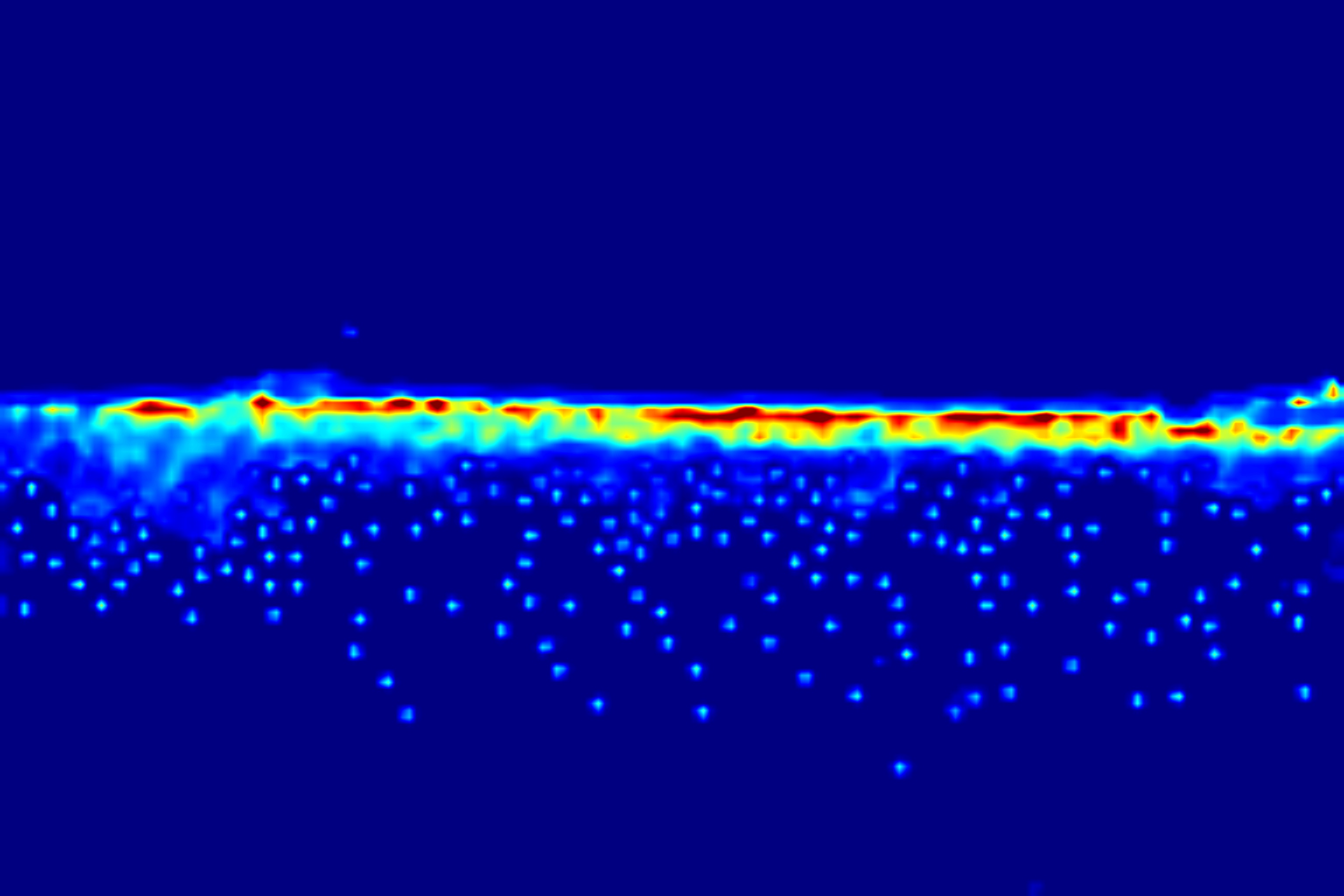}  &
			\includegraphics[height=0.13\linewidth]{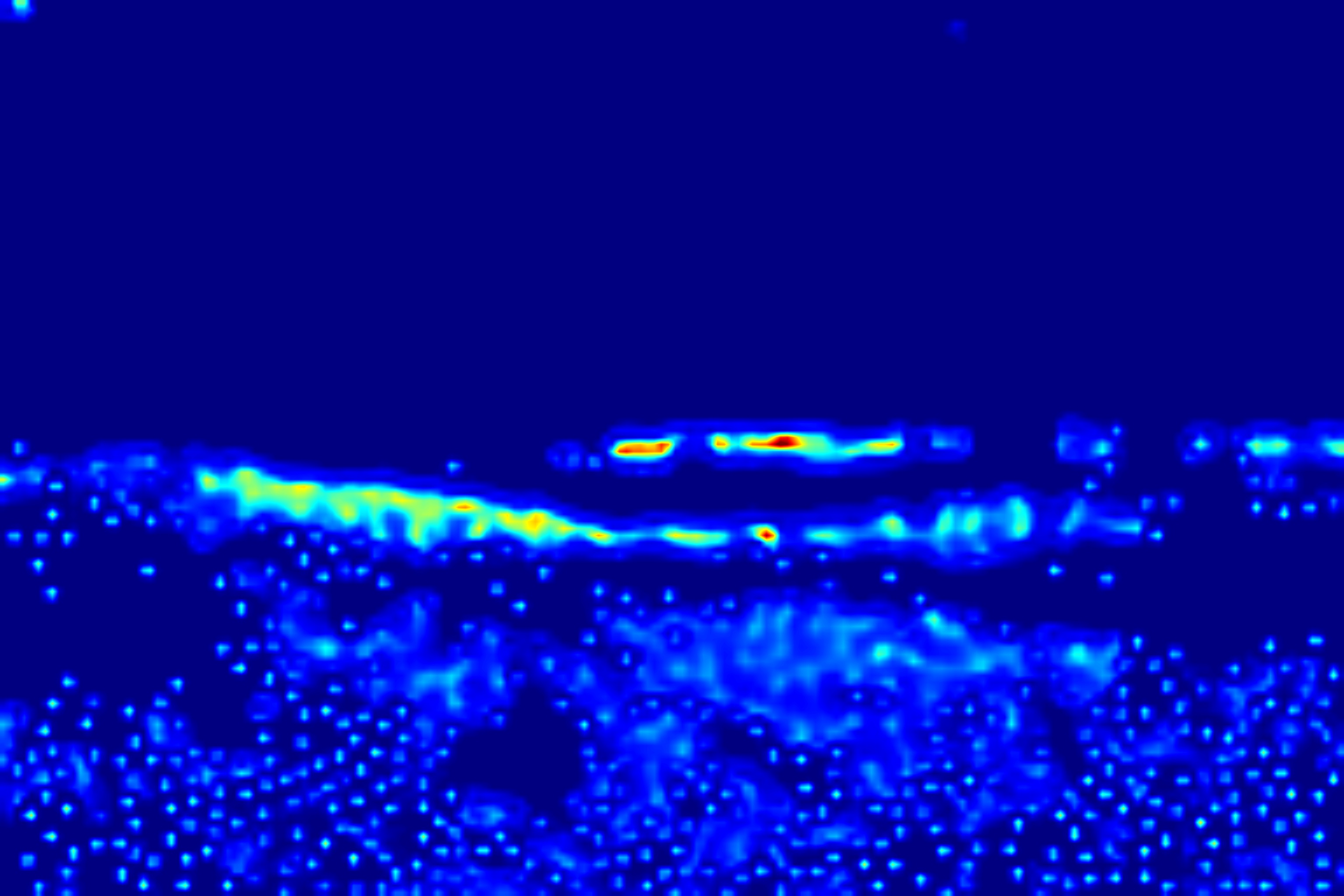}  &
			\includegraphics[height=0.13\linewidth]{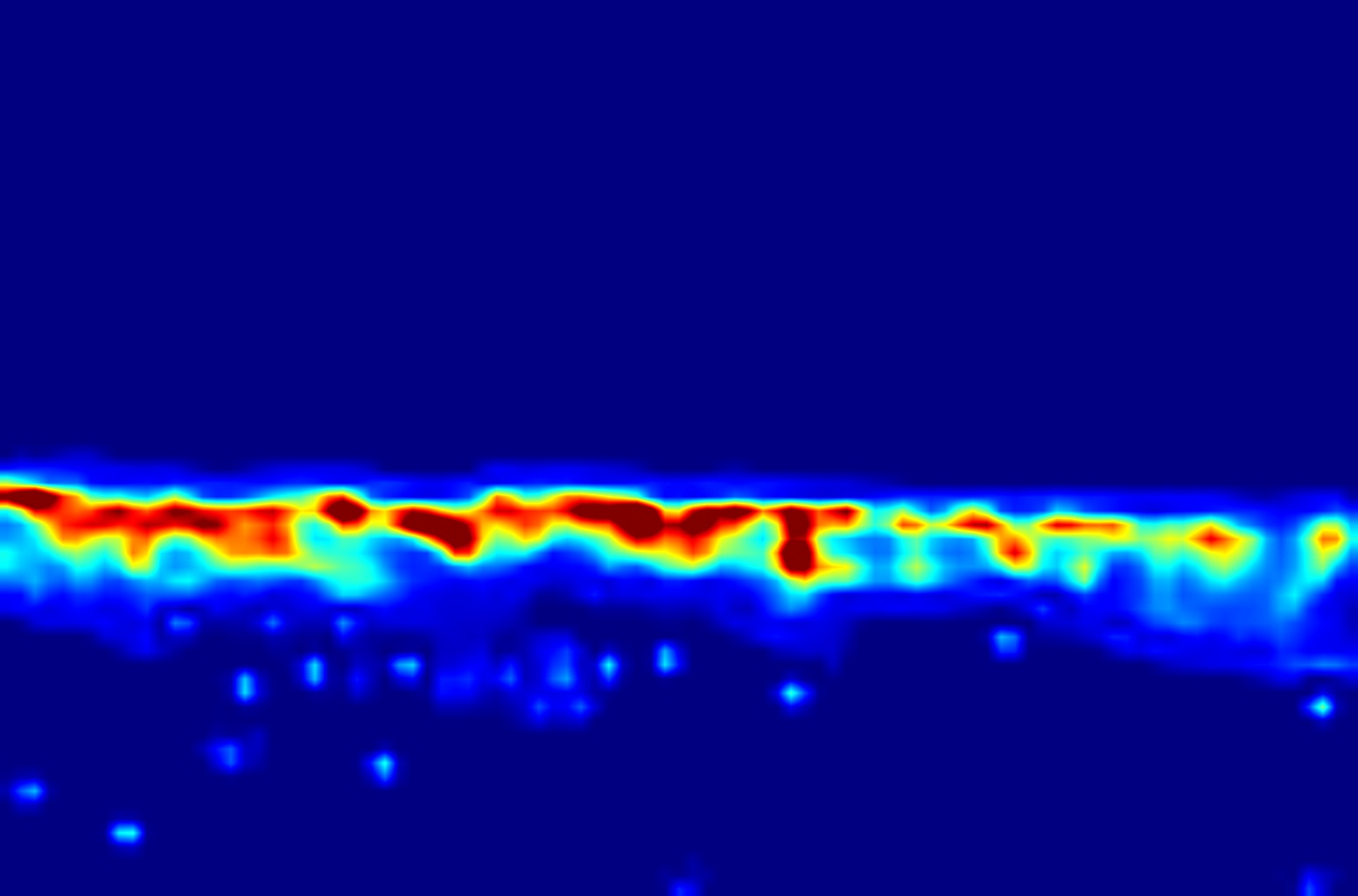} &
			\includegraphics[height=0.13\linewidth]{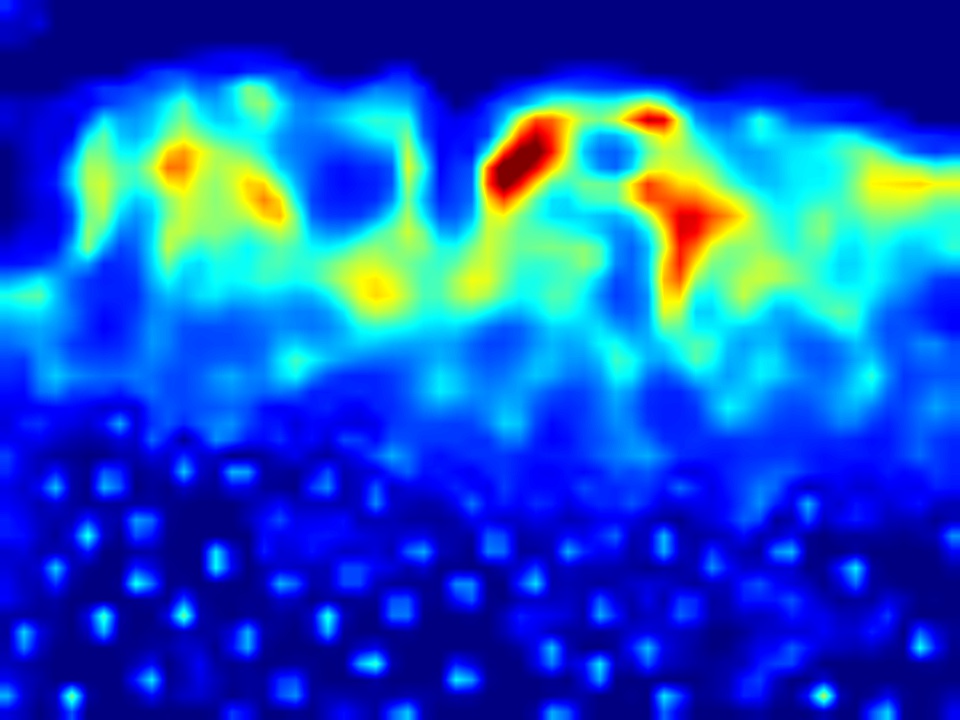} \\

			\footnotesize{\emmm{MAN}: 947.04} & \footnotesize{\emmm{MAN}: 1013.15} & \footnotesize{\emmm{MAN}: 1205.58} & \footnotesize{\emmm{MAN}: 404.02} & \footnotesize{\emmm{MAN}: 534.72} \\
		\end{tabular}
		\caption{Visualizations on UCF-QNRF. The first and third rows are input images while the second and forth rows are the corresponding density maps predicted by our MAN. The warmer color means higher density.}
		\label{fig:viz}
	\end{center}
\end{figure*}}

{\flushleft \textbf{Effect of $\delta$:}} We hold experiments to understand the parameter selection of proposed \emmm{Instance Attention Loss}. We compare the counting accuracy under different thresholds on UCF-QNRF, which result is shown in Figure~\ref{fig:ablation1}. 

We set \ourmodel without Instance Attention Loss as the baseline, which also means $\delta = 1$. We observe that the counting accuracy reaches best when $\delta = 0.9$, representing the model cuts off $10\%$ annotations with largest deviations from the prediction. 

As $\delta$ is selected smaller, the accuracy of \emmm{MAN} declines obviously. It can be explained by the insufficient use of ground truth and that the model is weakly supervised. Then, when we focus on $0.8 \leq \delta < 1$, the results are much better than supervision by all annotations. This may suggest that there are about $20\%$ annotations which will negatively influence the performance of model in counting when adopted in training. And by our \emmm{Instance Attention Loss}, it reduces this negative influence conveniently and effectively.

{\flushleft \textbf{Visualizations of LRA:}} Figure~\ref{fig:atten} presents a visualization of \XP{the} region mask $\widetilde{R}_i$ in Learnable Region Attention (LRA) where the location of corresponding feature $i$ is marked by a white circle. Supervised by the \emmm{Local Attention Regularization}, LRA is able to balance the allocations of attention resources. As can be seen, the attention region becomes gradually narrower as the scale of focus crowd is smaller and the number of people needed attention in each region is about the same. It is similar to human's efficient deployment of attention resources and justifies the usefulness of our LRA and LAR.

{\flushleft \textbf{Running Cost Evaluation:}} Table~\ref{tab:parameter} reports a comparison of model size, floating point operations (FLOPs) computed on \XP{one} $384 \times 384$ input image, inference time for $1024 \times 1024$ images. \XP{It can be easily observed that the model size and inference time of MAN are closed to those of VGG19+Trans and much smaller than those of ViT-B. Moreover, MAN and VGG19-Trans are with a marginal difference in FLOPs. It thus shows that the proposed components are lightweight compared with vanilla transformers.} 

\renewcommand{\tabcolsep}{3 pt}{
\begin{table}
\small
\begin{center}
\begin{tabular}{|c|cccc|}
\hline
 & ViT-B~\cite{dosovitskiy2020image} & Bayesian~\cite{ma2019bayesian} & VGG19+Trans & \textbf{MAN}\\
\hline
Model Size (M) & 86.0 & 21.5 & 29.9 & 30.9\\
GFLOPs & 55.4 & 56.9 & 58.0 & 58.2\\
Inference time & 21.3 & 10.3 & 10.8 & 11.3\\
\hline
\end{tabular}
\end{center}
\vspace{-4mm}
\caption{Comparison of the model size (M), FLOPs and Inference time (s / 100 images). Trans stands for the vanilla encoder. The computational cost of MAN only increases a little.}
\label{tab:parameter}
\vspace{-4mm}
\end{table}}
\section{Conclusion}

\XP{This paper is aimed to enhance the ability of transformers in spatial local context encoding for crowd counting. We contribute to the structure of transformers by proposing a Learnable Region Attention  module. We also improve the training pipeline by designing Local Attention Regularization to balance the attention allocated for each proposed region and introducing the Instance Attention Loss to reduce the influences of label noise.} The proposed \ourmodel has achieved state-of-the-art performances on four crowd counting datasets. We consider future directions for applying our model to a wider range of vision tasks.



{\small
\bibliographystyle{ieee_fullname}
\bibliography{egbib}

\begin{thebibliography}{10}\itemsep=-1pt

\bibitem{beltagy2020longformer}
Iz Beltagy, Matthew~E Peters, and Arman Cohan.
\newblock Longformer: The long-document transformer.
\newblock {\em arXiv preprint}, 2020.

\bibitem{cao2018scale}
Xinkun Cao, Zhipeng Wang, Yanyun Zhao, and Fei Su.
\newblock Scale aggregation network for accurate and efficient crowd counting.
\newblock In {\em ECCV}, 2018.

\bibitem{carion2020end}
Nicolas Carion, Francisco Massa, Gabriel Synnaeve, Nicolas Usunier, Alexander
  Kirillov, and Sergey Zagoruyko.
\newblock End-to-end object detection with transformers.
\newblock In {\em ECCV}, 2020.

\bibitem{chen2021transformer}
Xin Chen, Bin Yan, Jiawen Zhu, Dong Wang, Xiaoyun Yang, and Huchuan Lu.
\newblock Transformer tracking.
\newblock In {\em CVPR}, 2021.

\bibitem{collegio2019attention}
Andrew~J Collegio, Joseph~C Nah, Paul~S Scotti, and Sarah Shomstein.
\newblock Attention scales according to inferred real-world object size.
\newblock {\em Nature human behaviour}, 2019.

\bibitem{comon2014tensors}
Pierre Comon.
\newblock Tensors: a brief introduction.
\newblock {\em IEEE Signal Processing Magazine}, 31(3):44--53, 2014.

\bibitem{d2021convit}
St{\'e}phane d'Ascoli, Hugo Touvron, Matthew Leavitt, Ari Morcos, Giulio
  Biroli, and Levent Sagun.
\newblock Convit: Improving vision transformers with soft convolutional
  inductive biases.
\newblock {\em arXiv preprint arXiv:2103.10697}, 2021.

\bibitem{dehghani2018universal}
Mostafa Dehghani, Stephan Gouws, Oriol Vinyals, Jakob Uszkoreit, and Lukasz
  Kaiser.
\newblock Universal transformers.
\newblock In {\em ICLR}, 2018.

\bibitem{devlin2018bert}
Jacob Devlin, Ming-Wei Chang, Kenton Lee, and Kristina Toutanova.
\newblock Bert: Pre-training of deep bidirectional transformers for language
  understanding.
\newblock {\em arXiv preprint}, 2018.

\bibitem{dosovitskiy2020image}
Alexey Dosovitskiy, Lucas Beyer, Alexander Kolesnikov, Dirk Weissenborn,
  Xiaohua Zhai, Thomas Unterthiner, Mostafa Dehghani, Matthias Minderer, Georg
  Heigold, Sylvain Gelly, et~al.
\newblock An image is worth 16x16 words: Transformers for image recognition at
  scale.
\newblock In {\em ICLR}, 2020.

\bibitem{idrees2018composition}
Haroon Idrees, Muhmmad Tayyab, Kishan Athrey, Dong Zhang, Somaya Al-Maadeed,
  Nasir Rajpoot, and Mubarak Shah.
\newblock Composition loss for counting, density map estimation and
  localization in dense crowds.
\newblock In {\em ECCV}, 2018.

\bibitem{kingmaadam}
Diederik~P Kingma and Jimmy~Lei Ba.
\newblock Adam: Amethod for stochastic optimization.

\bibitem{li2018csrnet}
Yuhong Li, Xiaofan Zhang, and Deming Chen.
\newblock Csrnet: Dilated convolutional neural networks for understanding the
  highly congested scenes.
\newblock In {\em CVPR}, 2018.

\bibitem{lin2021direct}
Hui Lin, Xiaopeng Hong, Zhiheng Ma, Xing Wei, Yunfeng Qiu, Yaowei Wang, and
  Yihong Gong.
\newblock Direct measure matching for crowd counting.
\newblock In {\em IJCAI}, 2021.

\bibitem{liu2018decidenet}
Jiang Liu, Chenqiang Gao, Deyu Meng, and Alexander~G Hauptmann.
\newblock Decidenet: Counting varying density crowds through attention guided
  detection and density estimation.
\newblock In {\em CVPR}, 2018.

\bibitem{liu2019context}
Weizhe Liu, Mathieu Salzmann, and Pascal Fua.
\newblock Context-aware crowd counting.
\newblock In {\em CVPR}, 2019.

\bibitem{liu2020semi}
Yan Liu, Lingqiao Liu, Peng Wang, Pingping Zhang, and Yinjie Lei.
\newblock Semi-supervised crowd counting via self-training on surrogate tasks.
\newblock In {\em ECCV}, 2020.

\bibitem{liu2019point}
Yuting Liu, Miaojing Shi, Qijun Zhao, and Xiaofang Wang.
\newblock Point in, box out: Beyond counting persons in crowds.
\newblock In {\em CVPR}, 2019.

\bibitem{liu2021swin}
Ze Liu, Yutong Lin, Yue Cao, Han Hu, Yixuan Wei, Zheng Zhang, Stephen Lin, and
  Baining Guo.
\newblock Swin transformer: Hierarchical vision transformer using shifted
  windows.
\newblock {\em arXiv preprint}, 2021.

\bibitem{ma2021towards}
Zhiheng Ma, Xiaopeng Hong, Xing Wei, Yunfeng Qiu, and Yihong Gong.
\newblock Towards a universal model for cross-dataset crowd counting.
\newblock In {\em ICCV}, 2021.

\bibitem{ma2019bayesian}
Zhiheng Ma, Xing Wei, Xiaopeng Hong, and Yihong Gong.
\newblock Bayesian loss for crowd count estimation with point supervision.
\newblock In {\em ICCV}, 2019.

\bibitem{ma2020learning}
Zhiheng Ma, Xing Wei, Xiaopeng Hong, and Yihong Gong.
\newblock Learning scales from points: A scale-aware probabilistic model for
  crowd counting.
\newblock In {\em ACMMM}, 2020.

\bibitem{ma2021learning}
Zhiheng Ma, Xing Wei, Xiaopeng Hong, Hui Lin, Yunfeng Qiu, and Yihong Gong.
\newblock Learning to count via unbalanced optimal transport.
\newblock In {\em AAAI}, 2021.

\bibitem{naskovska2020using}
Kristina Naskovska, André L.~F. de Almeida, and Martin Haardt.
\newblock Using double contractions to derive the structure of slice-wise
  multiplications of tensors with applications to semi-blind mimo ofdm, 2020.

\bibitem{nguyen2020differentiable}
Thanh-Tung Nguyen, Xuan-Phi Nguyen, Shafiq Joty, and Xiaoli Li.
\newblock Differentiable window for dynamic local attention.
\newblock {\em arXiv preprint arXiv:2006.13561}, 2020.

\bibitem{radfordimproving}
Alec Radford, Karthik Narasimhan, Tim Salimans, and Ilya Sutskever.
\newblock Improving language understanding by generative pre-training.
\newblock 2018.

\bibitem{shen2018disan}
Tao Shen, Tianyi Zhou, Guodong Long, Jing Jiang, Shirui Pan, and Chengqi Zhang.
\newblock Disan: Directional self-attention network for rnn/cnn-free language
  understanding.
\newblock In {\em AAAI}, 2018.

\bibitem{simonyan2014very}
Karen Simonyan and Andrew Zisserman.
\newblock Very deep convolutional networks for large-scale image recognition.
\newblock {\em arXiv preprint}, 2014.

\bibitem{sindagi2020jhu}
Vishwanath Sindagi, Rajeev Yasarla, and Vishal~MM Patel.
\newblock Jhu-crowd++: Large-scale crowd counting dataset and a benchmark
  method.
\newblock {\em PAMI}, 2020.

\bibitem{sindagi2017generating}
Vishwanath~A Sindagi and Vishal~M Patel.
\newblock Generating high-quality crowd density maps using contextual pyramid
  cnns.
\newblock In {\em ICCV}, 2017.

\bibitem{song2021rethinking}
Qingyu Song, Changan Wang, Zhengkai Jiang, Yabiao Wang, Ying Tai, Chengjie
  Wang, Jilin Li, Feiyue Huang, and Yang Wu.
\newblock Rethinking counting and localization in crowds: A purely point-based
  framework.
\newblock In {\em ICCV}, 2021.

\bibitem{strudel2021segmenter}
Robin Strudel, Ricardo Garcia, Ivan Laptev, and Cordelia Schmid.
\newblock Segmenter: Transformer for semantic segmentation.
\newblock {\em arXiv preprint}, 2021.

\bibitem{sun2020transtrack}
Peize Sun, Yi Jiang, Rufeng Zhang, Enze Xie, Jinkun Cao, Xinting Hu, Tao Kong,
  Zehuan Yuan, Changhu Wang, and Ping Luo.
\newblock Transtrack: Multiple-object tracking with transformer.
\newblock {\em arXiv preprint}, 2020.

\bibitem{sun2021rethinking}
Zhiqing Sun, Shengcao Cao, Yiming Yang, and Kris~M Kitani.
\newblock Rethinking transformer-based set prediction for object detection.
\newblock In {\em ICCV}, 2021.

\bibitem{vaswani2021scaling}
Ashish Vaswani, Prajit Ramachandran, Aravind Srinivas, Niki Parmar, Blake
  Hechtman, and Jonathon Shlens.
\newblock Scaling local self-attention for parameter efficient visual
  backbones.
\newblock In {\em CVPR}, 2021.

\bibitem{vaswani2017attention}
Ashish Vaswani, Noam Shazeer, Niki Parmar, Jakob Uszkoreit, Llion Jones,
  Aidan~N Gomez, {\L}ukasz Kaiser, and Illia Polosukhin.
\newblock Attention is all you need.
\newblock In {\em Advances in neural information processing systems}, 2017.

\bibitem{wan2021generalized}
Jia Wan, Ziquan Liu, and Antoni~B Chan.
\newblock A generalized loss function for crowd counting and localization.
\newblock In {\em CVPR}, 2021.

\bibitem{wang2020distribution}
Boyu Wang, Huidong Liu, Dimitris Samaras, and Minh~Hoai Nguyen.
\newblock Distribution matching for crowd counting.
\newblock {\em NIPS}, 2020.

\bibitem{wang2021transformer}
Ning Wang, Wengang Zhou, Jie Wang, and Houqiang Li.
\newblock Transformer meets tracker: Exploiting temporal context for robust
  visual tracking.
\newblock In {\em CVPR}, 2021.

\bibitem{wang2020nwpu}
Qi Wang, Junyu Gao, Wei Lin, and Xuelong Li.
\newblock Nwpu-crowd: A large-scale benchmark for crowd counting and
  localization.
\newblock {\em PAMI}, 2020.

\bibitem{wang2022eccnas}
Yabin Wang, Zhiheng Ma, Xing Wei, Shuai Zheng, Yaowei Wang, and Xiaopeng Hong.
\newblock Eccnas: Efficient crowd counting neural architecture search.
\newblock {\em ACM TOMM}, 2022.

\bibitem{wang2021end}
Yuqing Wang, Zhaoliang Xu, Xinlong Wang, Chunhua Shen, Baoshan Cheng, Hao Shen,
  and Huaxia Xia.
\newblock End-to-end video instance segmentation with transformers.
\newblock In {\em CVPR}, 2021.

\bibitem{xu2021vitae}
Yufei Xu, Qiming Zhang, Jing Zhang, and Dacheng Tao.
\newblock Vitae: Vision transformer advanced by exploring intrinsic inductive
  bias.
\newblock {\em arXiv preprint}, 2021.

\bibitem{yan2019perspective}
Zhaoyi Yan, Yuchen Yuan, Wangmeng Zuo, Xiao Tan, Yezhen Wang, Shilei Wen, and
  Errui Ding.
\newblock Perspective-guided convolution networks for crowd counting.
\newblock In {\em ICCV}, 2019.

\bibitem{yang2021focal}
Jianwei Yang, Chunyuan Li, Pengchuan Zhang, Xiyang Dai, Bin Xiao, Lu Yuan, and
  Jianfeng Gao.
\newblock Focal self-attention for local-global interactions in vision
  transformers.
\newblock {\em arXiv preprint}, 2021.

\bibitem{yang2020reverse}
Yifan Yang, Guorong Li, Zhe Wu, Li Su, Qingming Huang, and Nicu Sebe.
\newblock Reverse perspective network for perspective-aware object counting.
\newblock In {\em CVPR}, 2020.

\bibitem{yoon2018dynamic}
Deunsol Yoon, Dongbok Lee, and SangKeun Lee.
\newblock Dynamic self-attention: Computing attention over words dynamically
  for sentence embedding.
\newblock {\em arXiv preprint}, 2018.

\bibitem{zeng2017multi}
Lingke Zeng, Xiangmin Xu, Bolun Cai, Suo Qiu, and Tong Zhang.
\newblock Multi-scale convolutional neural networks for crowd counting.
\newblock In {\em ICIP}, 2017.

\bibitem{zhang2021multi}
Pengchuan Zhang, Xiyang Dai, Jianwei Yang, Bin Xiao, Lu Yuan, Lei Zhang, and
  Jianfeng Gao.
\newblock Multi-scale vision longformer: A new vision transformer for
  high-resolution image encoding.
\newblock {\em arXiv preprint}, 2021.

\bibitem{zhang2016single}
Yingying Zhang, Desen Zhou, Siqin Chen, Shenghua Gao, and Yi Ma.
\newblock Single-image crowd counting via multi-column convolutional neural
  network.
\newblock In {\em CVPR}, 2016.

\bibitem{zhao2019leveraging}
Muming Zhao, Jian Zhang, Chongyang Zhang, and Wenjun Zhang.
\newblock Leveraging heterogeneous auxiliary tasks to assist crowd counting.
\newblock In {\em CVPR}, 2019.

\bibitem{zheng2020end}
Minghang Zheng, Peng Gao, Xiaogang Wang, Hongsheng Li, and Hao Dong.
\newblock End-to-end object detection with adaptive clustering transformer.
\newblock {\em arXiv preprint}, 2020.

\bibitem{zheng2021rethinking}
Sixiao Zheng, Jiachen Lu, Hengshuang Zhao, Xiatian Zhu, Zekun Luo, Yabiao Wang,
  Yanwei Fu, Jianfeng Feng, Tao Xiang, Philip~HS Torr, et~al.
\newblock Rethinking semantic segmentation from a sequence-to-sequence
  perspective with transformers.
\newblock In {\em CVPR}, 2021.

\bibitem{zhu2020deformable}
Xizhou Zhu, Weijie Su, Lewei Lu, Bin Li, Xiaogang Wang, and Jifeng Dai.
\newblock Deformable detr: Deformable transformers for end-to-end object
  detection.
\newblock In {\em ICLR}, 2020.

\end{thebibliography}
}

\end{document}